%% file: arxiv_main.tex
\documentclass[runningheads]{llncs}
\usepackage{graphicx}

\usepackage{tikz}
\usepackage{comment}
\usepackage{amsmath,amssymb} %
\usepackage{color}

\usepackage[accsupp]{axessibility}  %

\usepackage[pagebackref,breaklinks,colorlinks]{hyperref}
\usepackage{graphicx}
\usepackage{amsmath}
\usepackage{xspace}
\usepackage{textcomp}
\usepackage[export]{adjustbox}
\usepackage{rotating}
\usepackage{wrapfig}
\usepackage{float}
\usepackage[capitalize]{cleveref}
\usepackage{siunitx}
\usepackage{array}
\newcolumntype{P}[1]{>{\centering\arraybackslash}p{#1}}
\usepackage{booktabs}
\usepackage{cite}
\usepackage{subfig}
\usepackage{pgfplots}
\pgfplotsset{compat=1.16}

\begin{document}
\pagestyle{headings}
\mainmatter
\def\ECCVSubNumber{5144}  %

\title{GAN Cocktail: mixing GANs without dataset access} %

\titlerunning{ECCV-22 submission ID \ECCVSubNumber} 
\authorrunning{ECCV-22 submission ID \ECCVSubNumber} 
\author{Anonymous ECCV submission}
\institute{Paper ID \ECCVSubNumber}

\titlerunning{GAN Cocktail}

\author{Omri Avrahami\inst{1} \and
Dani Lischinski\inst{1} \and
Ohad Fried\inst{2}}

\authorrunning{Avrahami et al.}

\institute{$^1$The Hebrew University of Jerusalem \hspace{1cm}
$^2$Reichman University}

\maketitle

\def\ShowNotes{}
\input{macros.tex}

\input{sections/abstract}
\input{sections/introduction}
\input{sections/related_work}
\input{sections/method}
\input{sections/results}
\input{sections/limitations}
\input{sections/social_implications}
\input{sections/conclusions}

\smallskip
\textbf{Acknowledgments} This work was supported in part by Lightricks Ltd and by the Israel Science Foundation (grants No. 2492/20, 1574/21, and 2611/21).

\bibliographystyle{splncs04}
\bibliography{egbib}

\clearpage
\appendix
\input{appendices/implementation_details.tex}
\input{appendices/additional_experiments.tex}
\input{appendices/datasets.tex}

\input{appendices/training.tex}
\input{appendices/applications.tex}
\input{appendices/uncurated_examples.tex}

\end{document}

%% file: macros.tex
\newcommand{\ignorethis}[1]{}
\newcommand{\redund}[1]{#1}

\newcommand{\etal       }     {{et~al.}}
\newcommand{\apriori    }     {\textit{a~priori}}
\newcommand{\aposteriori}     {\textit{a~posteriori}}
\newcommand{\perse      }     {\textit{per~se}}
\newcommand{\naive      }     {{na\"{\i}ve}}

\newcommand{\Identity   }     {\mat{I}}
\newcommand{\Zero       }     {\mathbf{0}}
\newcommand{\Reals      }     {{\textrm{I\kern-0.18em R}}}
\newcommand{\isdefined  }     {\mbox{\hspace{0.5ex}:=\hspace{0.5ex}}}
\newcommand{\texthalf   }     {\ensuremath{\textstyle\frac{1}{2}}}
\newcommand{\half       }     {\ensuremath{\frac{1}{2}}}
\newcommand{\third      }     {\ensuremath{\frac{1}{3}}}
\newcommand{\fourth     }     {\ensuremath{\frac{1}{4}}}

\newcommand{\Lone} {\ensuremath{L_1}}
\newcommand{\Ltwo} {\ensuremath{L_2}}

\newcommand{\mat        } [1] {{\text{\boldmath $\mathbit{#1}$}}}
\newcommand{\Approx     } [1] {\widetilde{#1}}
\newcommand{\change     } [1] {\mbox{{\footnotesize $\Delta$} \kern-3pt}#1}

\newcommand{\Order      } [1] {O(#1)}
\newcommand{\set        } [1] {{\lbrace #1 \rbrace}}
\newcommand{\floor      } [1] {{\lfloor #1 \rfloor}}
\newcommand{\ceil       } [1] {{\lceil  #1 \rceil }}
\newcommand{\inverse    } [1] {{#1}^{-1}}
\newcommand{\transpose  } [1] {{#1}^\mathrm{T}}
\newcommand{\invtransp  } [1] {{#1}^{-\mathrm{T}}}
\newcommand{\relu       } [1] {{\lbrack #1 \rbrack_+}}

\newcommand{\abs        } [1] {{| #1 |}}
\newcommand{\Abs        } [1] {{\left| #1 \right|}}
\newcommand{\norm       } [1] {{\| #1 \|}}
\newcommand{\Norm       } [1] {{\left\| #1 \right\|}}
\newcommand{\pnorm      } [2] {\norm{#1}_{#2}}
\newcommand{\Pnorm      } [2] {\Norm{#1}_{#2}}
\newcommand{\inner      } [2] {{\langle {#1} \, | \, {#2} \rangle}}
\newcommand{\Inner      } [2] {{\left\langle \begin{array}{@{}c|c@{}}
                               \displaystyle {#1} & \displaystyle {#2}
                               \end{array} \right\rangle}}

\newcommand{\twopartdef}[4]
{
  \left\{
  \begin{array}{ll}
    #1 & \mbox{if } #2 \\
    #3 & \mbox{if } #4
  \end{array}
  \right.
}

\newcommand{\fourpartdef}[8]
{
  \left\{
  \begin{array}{ll}
    #1 & \mbox{if } #2 \\
    #3 & \mbox{if } #4 \\
    #5 & \mbox{if } #6 \\
    #7 & \mbox{if } #8
  \end{array}
  \right.
}

\newcommand{\len}[1]{\text{len}(#1)}

\newlength{\w}
\newlength{\h}
\newlength{\x}

\definecolor{darkred}{rgb}{0.7,0.1,0.1}
\definecolor{darkgreen}{rgb}{0.1,0.6,0.1}
\definecolor{cyan}{rgb}{0.7,0.0,0.7}
\definecolor{otherblue}{rgb}{0.1,0.4,0.8}
\definecolor{maroon}{rgb}{0.76,.13,.28}
\definecolor{burntorange}{rgb}{0.81,.33,0}

\ifdefined\ShowNotes
  \newcommand{\colornote}[3]{{\color{#1}\textbf{#2} #3\normalfont}}
\else
  \newcommand{\colornote}[3]{}
\fi

\newcommand {\todo}[1]{\colornote{cyan}{TODO}{#1}}
\newcommand {\ohad}[1]{\colornote{otherblue}{OF:}{#1}}
\newcommand {\dani}[1]{\colornote{darkgreen}{DL:}{#1}}
\newcommand {\omri}[1]{\colornote{burntorange}{OA:}{#1}}

\newcommand {\reqs}[1]{\colornote{red}{\tiny #1}}

\newcommand {\new}[1]{{\color{red}{#1}}}

\newcommand*\rot[1]{\rotatebox{90}{#1}}

\newcommand {\newstuff}[1]{#1}

\newcommand\todosilent[1]{}

\newcommand{\woBGmask}{{w/o~bg~\&~mask}}
\newcommand{\woMask}{{w/o~mask}}

\providecommand{\keywords}[1]
{
  \textbf{\textit{Keywords---}} #1
}

\newcommand {\shortcite}[1]{\cite{#1}}

\newcommand{\GAN}{\textit{GAN}}
\newcommand{\data}{\mathit{data}}
\newcommand{\unionGAN}{\textsc{UnionGAN}\xspace}
\newcommand {\ganArrow}[2]{\ensuremath{\GAN_{{#1} \rightarrow {#2}}}}
\newcommand {\gan}[1]{\ensuremath{\GAN_{#1}}}

%% file: sections/abstract.tex
\begin{abstract}
Today's generative models are capable of synthesizing high-fidelity images, but each model specializes on a specific target domain. This raises the need for model merging: combining two or more pretrained generative models into a single unified one. In this work we tackle the problem of model merging, given two constraints that often come up in the real world: (1) no access to the original training data, and (2) without increasing the network size. To the best of our knowledge, model merging under these constraints has not been studied thus far. 
We propose a novel, two-stage solution\footnote{Code is available at: \url{https://omriavrahami.com/GAN-cocktail-page/}}. In the first stage, we transform the weights of all the models to the same parameter space by a technique we term model rooting. In the second stage, we merge the rooted models by averaging their weights and fine-tuning them for each specific domain, using only data generated by the original trained models. We demonstrate that our approach is superior to baseline methods and to existing transfer learning techniques, and investigate several applications.

\keywords{Generative Adversarial Networks, Model Merging}
\end{abstract}

%% file: sections/introduction.tex
\section{Introduction}
\label{sec:introduction}

Generative adversarial networks (GANs) \cite{gan} have achieved impressive results in neural image synthesis \cite{biggan, karras2017progressive, stylegan1, stylegan2}.
However, these generative models typically specialize on a specific image domain, such as human faces, kitchens, or landscapes. This is in contrary to traditional computer graphics, where a general purpose representation (e.g., textured meshes) and a general purpose renderer can produce images of diverse object types and scenes. In order to extend the applicability and versatility of neural image synthesis, in this work we explore \textit{model merging} --- the process of combining two or more generative models into a single conditional model.
There are several concrete benefits to model merging:
\begin{enumerate}
    \item It is well suited for decentralized workflows. Different entities can collect their own datasets and train their own models, which may later be merged.
    \item If performed properly, merged models can reduce memory and computation requirements, enabling their use on edge devices with limited resources.
    \item Merged models enable semantic editing across domains, as described next.
\end{enumerate}

GANs often produce a semantically meaningful latent space.
Several embedding techniques \cite{image2stylegan, image2stylegan++, pidhorskyi2020adversarial} have been proposed to map real input images to latent codes of a pre-trained GAN generator, which enables semantic manipulation. Images can be interpolated and transformed using semantic vectors in the embedding space \cite{harkonen2020ganspace,shen2020interpreting}, effectively using it as a strong regularizer. 
A problem arises when one wants to use several pre-trained generators for semantic manipulations (e.g., interpolating between images from GAN~$A$ and GAN~$B$) --- the different models do not share the same latent representation, and hence do not ``speak the same language''. 
Model merging places several GANs in a shared latent space, allowing such cross-domain semantic manipulations.

We tackle the problem of merging several GAN models into a single one under the following real-world constraints:
\begin{enumerate}
    \item \textbf{No access to training data.} Many GAN models are being released without the data that they were trained on. This can occur because datasets are too large \cite{chen2020generative, radford2021learning, ramesh2021zero} or due to privacy/copyright issues.
    Hence, we assume that no training data is available, and only rely on data generated by the pre-trained models. 
    \item \textbf{Limited computing power.} A \naive{} approach to merging several GAN models is to sample from them separately (e.g., by multinomial sampling functions \cite{wang2020minegan}). With this approach, the model size and inference time grow linearly with the number of GAN models, which may not be practical due to lack of computing power (e.g., edge devices). 
    In addition, this approach does not result in a shared latent space, so it does not support cross-domain semantic manipulations as described earlier. Our goal is to maintain a constant network size, regardless of the number of GANs being merged.
\end{enumerate}

To the best of our knowledge, performing model merging under these constraints has not been studied yet.
This is a challenging task:
pre-trained GANs typically do not model the entire real image distribution \cite{bau2019seeing}; hence, learning from the outputs of pre-trained models will be sub-optimal. In addition, the constraint on the model size may reduce its capacity. 

We start by adapting existing solutions from the field of transfer-learning as baselines (\Cref{sec:baselines}). Next, we present our novel two-stage solution for model merging.
We first transfer the weights of the input models to a joint semantic space using a technique we term model rooting (\Cref{subsec:rooting}). We then merge the rooted models via averaging and fine-tuning (\Cref{subsec:finetune}). We find that model rooting introduces an inductive bias that helps the merged model achieve superior results compared to baselines and to existing transfer-learning methods (\Cref{sec:results}).

To summarize, this paper has the following contributions:
\begin{itemize}
    \item We introduce the real-world problem of merging several GAN models without access to their training data and with no increase in model size or inference time.
    \item We adapt several transfer-learning techniques to the GAN merging problem.
    \item We introduce a novel two-stage approach for GAN merging and evaluate its performance.
\end{itemize}

%% file: sections/related_work.tex
\section{Related Work}
\label{sec:related_work}

\textbf{Generative adversarial networks:} GANs \cite{gan} consist of a generator $G$ and a discriminator $D$ that compete in a two-player minimax game: the discriminator tries to distinguish real training data from generated data, and the generator tries to fool the discriminator. Training GANs is difficult, due to mode collapse and training instability, and several methods focus on addressing these problems \cite{gulrajani2017improved, mao2017least, mescheder2018training, miyato2018spectral, salimans2016improved}, while another line of research aims to improve the architectures to generate higher quality images \cite{biggan, karras2017progressive, stylegan1, stylegan2}. Karras \etal~\cite{stylegan1, stylegan2} introduced the StyleGAN architecture that leads to an automatically learned, unsupervised separation of high-level
attributes and stochastic variation in the generated images and enables intuitive, scale-specific control over the synthesis. For our experiments we use the StyleGAN2 framework.
It is important to note that our approach, as well as the baselines, are model-agnostic and there is no dependency on any StyleGAN-specific capabilities. We demonstrate mixing between models of different architectures in
Supp.~Section 2.1.

\textbf{Transfer learning:} Learning how to transfer knowledge from a source domain to a target domain is a well-studied problem in machine learning \cite{pan2009survey, oquab2014learning, donahue2014decaf,kornblith2019better, tran2019transferability, nguyen2020leep, bao2019information}, mainly in the discriminative case. It is important to note that the transfer-learning literature focuses on the case where there is a training dataset for the target domain, which is not the case in our scenario. Recent works that are more related to our problem by Shu et al. \cite{shu2021zoo} and Geyer et al. \cite{geyer2019transfer} demonstrate the ability to perform transfer learning from several source models into a single target model. However they are not applicable to our setting because: (1) neither method tackles generative models, as we do; (2) both of these methods assume that all the source models share the same architecture, whereas our problem formulation specifically focuses on the general case with arbitrary architectures (which is the real-world scenario, especially for generative models); (3) both methods assume access to training data, while we assume that the training data is unavailable; (4) T-IMM method \cite{geyer2019transfer} assumes that the user trains the \emph{source models} incrementally, which is different from our setting, where the source models training is not under the user control. To conclude, the current literature mainly focuses on the discriminative case and assumes access to the training data.

As shown by Wang \etal~\cite{transferring_gans}, the principles of transfer learning can be applied to image generation with GANs. Later, Noguchi and Harada \cite{noguchi2019image} proposed to constrain the training process to only update the normalization parameters instead of all of the model's trainable parameters. This shrinks the model capacity, which mitigates the overfitting problem and enables fine-tuning with an extremely small dataset.
However, limiting the capacity of the model enables to only change the style of the objects but not their shape, which isn't applicable to our setting, where the merged image domains may exhibit objects of completely different shapes.

Another approach for GAN transfer learning consists of adding a layer that steers the generated distribution towards the target distribution, which is also applicable for sampling from several models~\cite{wang2020minegan}. However, this approach stitches the source models together, and thus the model size and the inference time grow linearly in the number of source models. In addition, the models in this approach do not share the same latent space which limits their applicability.

\textbf{Continual learning:}
Continual Learning, also known as lifelong learning, is a setting where a model learns a large number of tasks sequentially without forgetting knowledge obtained from the preceding tasks, even though their training data is no longer available while training newer ones. Continual learning mainly deals with the ``catastrophic forgetting'' phenomenon, i.e., learning consecutive tasks without forgetting how to perform previously trained ones. Most previous efforts focused on classification tasks \cite{kirkpatrick2017overcoming, li2017learning, zenke2017continual}, and were later also adapted to the generative case \cite{seff2017continual}. 

Again, the literature focuses on the case where the new dataset is available, while the old dataset is not, which is not the case in our scenario, but we can adapt it to our setting, and do so in \Cref{sec:ewc_baseline}. Also, approaches that rely on designated architectures (e.g., lifelong GANs \cite{zhai2019lifelong}) cannot be adapted to our setting.

Another approach addressing the catastrophic forgetting problem that was proposed by Wu \etal~\cite{wu2018memory} is a memory-replay mechanism that uses the old generative model as a proxy to the old data. Our adaptation of the method of Wang \etal~\cite{transferring_gans} to our setting in \Cref{sec:transfer_gan} may be viewed also as adapting the approach of Wu \etal~\cite{wu2018memory}.

%% file: sections/method.tex
\section{Problem Formulation and Baselines}
\label{sec:baselines}

Our goal is to merge several GANs without access to their original training data. For example, given two trained GAN models, one that generates images of cats and another that generates images of dogs, we want to train a new single GAN model that produces images from both domains, without increasing the model size.
Below, we first formulate the problem, present several baselines, and then introduce our approach to solving this task in Section \ref{sec:our-approach}.

\subsection{Problem Formulation}

We are given $N$ GANs: $\{\gan{i} = (G_i, D_i)\}_{i=1}^{N}$, where $\gan{i}$ is pretrained on dataset $\data_i$ and consists of a generator $G_i$ and a discriminator $D_i$.
We denote the distribution of images that are produced by the generator $G_i$ by $P_{G_i}(z)$ and the real data distribution as $P_{{\data}_i}(x)$. 

Our goal is to create a ``union GAN'', $\unionGAN = (G_u, D_u)$, which is a conditional GAN \cite{cgan}, with the condition $c$ indicating which of the $N$ domains the generated sample should come from:
$\forall_{c \in [N]} P_{G_u}(z, c) = P_{{data}_c}(x)$ and $P_{D_u}(x, c)$ is the probability that $x$ came from ${data}_c$, rather than $P_{G_u}$. Note that the datasets $\data_i$ are not provided. Furthermore, the $N$ pretrained GAN models may have different architectures. Below, we adapt some current techniques from the transfer learning literature to address this problem.

\subsection{Baseline A: Training From Scratch}
\label{scratch}

Arguably, the simplest approach would be to train $\unionGAN$ from scratch, by using the samples generated by the pretrained input GANs as the only training data.
The objective function of a two-player minimax game that we aim to solve in this case is:
\begin{equation}
\label{cgan_equation}
\begin{split}
  \min_G \max_D V(D, G) = \mathbb{E}_{c \in [N], z \sim p_{z}(z), x \sim P_{G_c}(z)}[\log D(x, c)] +  \\
  \mathbb{E}_{c \in [N], z \sim p_{z}(z)}[\log (1 - D(G(z, c), c))]  
\end{split}
\end{equation}

Note that this formulation differs from a standard GAN in two ways: the discriminator is trained on the outputs of the given generators instead of on real data, and $\unionGAN$ is conditioned on both the class and the latent code~$z$. Thus, we simply treat the pre-trained generators as procedural sources of training data. We convert the unconditional model into a conditional one by adding a class embedding layer to the generator and concatenate its output the the latent code~$z$. An embedding layer is also added to the discriminator. See the supplementary material for more details.

Although the number of generated images that can be produced by a generator is unlimited (in contrast to a real training dataset),
we found that training using the real dataset produces better results.
This is likely caused by the fact that the pretrained GANs generate only a subset of the training data manifold.
To validate that the issue is not due to limited capacity, we train $\unionGAN$ on the degenerate case of $N = 1$, using different pretrained GANs, and observe a consistent increase in the resulting FID score, as reported in Table~\ref{tab:real_vs_synth}. In general, the best results can be achieved using the original real training data. 

\input{tables/data_vs_synth}

\subsection{Baseline B: TransferGAN}
\label{sec:transfer_gan}

The above method uses only the outputs of the pretrained models, thereby using them as black boxes.
Below we improve this method by using not only the generated data, but also the weights of the trained models.

Specifically, we adapt TransferGAN \cite{transferring_gans} to our problem as follows:
we initialize the \unionGAN with the $i$-th source model, and then train it on the outputs of all the GAN models (as described in the previous section) until convergence. Thus, we treat one of the models as both an initializer and a data source, and the remaining models as training data sources.

Compared to training from scratch, such initialization lowers the total FID score (for the union of the datasets), as reported in \Cref{tab:full_fid_results}. Furthermore, \Cref{tab:per_class_fid_comparison} shows that the FID score is lowered not only for the $i$-th dataset, but for the other datasets as well.

\subsection{Baseline C: Elastic Weight Consolidation}
\label{sec:ewc_baseline}

We observe that although the TransferGAN approach improves the final FID score, if we focus on the source model, we can see that its FID score (on the source class) is initially high and is degraded over the training process (the FID score of the original dataset class increases while the FIDs for the other classes decrease). This occurs due to catastrophic forgetting \cite{kirkpatrick2017overcoming} and can be mitigated by Elastic Weight Consolidation (EWC) \cite{kirkpatrick2017overcoming, gan_ewc}, applied to TransferGAN.

In order to assess the importance of the model parameters to its accuracy, we use Fisher information, which formulates how well we estimate the model parameters given the observations. In order to compute the empirical Fisher information given a pretrained model for a parameter $\theta_i$, we generate a certain amount of data $X$ and compute: 
$F_i = \mathbb{E} [(\frac{\partial}{\partial {\theta_i}} \mathcal{L}(x | {\theta_i}))^2]$
where $\mathcal{L}(x | {\theta_i})$ is the log-likelihood. In the generative case, we can equivalently compute the binary cross-entropy loss using the outputs of the discriminator that is fed by the outputs of the generator.

Thus, feeding the discriminator with the generator's outputs, we generate a large number of random samples, compute the binary cross-entropy loss on them and compute the derivative via back-propagation. We can add to our loss term the Elastic Weight Consolidation (EWC) penalty: 
$\mathcal{L}_{\textit{EWC}} = \mathcal{L}_{adv} + \lambda \sum_i{F_i(\theta_i - \theta_{S,i})}$
where $\theta_{S}$ represents the weights learned from the source domain, $i$ is the index of each parameter of
the model and $\lambda$ is the regularization weight to balance different losses.

Unfortunately, as can be seen in \Cref{tab:per_class_fid_comparison}, this procedure mitigates the catastrophic forgetting phenomena at the expense of degrading performance on the other classes.
We also experimented with a more \naive{} approach of applying a L2 loss on the source model weights, but, as we expected, the results were much worse for all but the source class.

\section{Our Approach: GAN Cocktail}
\label{sec:our-approach}

The main limitation of the transfer-learning approach is that it only uses the weights of one of the pre-trained GAN models ($\gan{i}$, the source model). In order to leverage the weights of all models, we propose a two-stage approach: At the first stage we perform \emph{model rooting} for all the input GAN models and in the second stage we perform \emph{model merging} by averaging the weights of the rooted models and then fine-tuning them using only data generated by the original models to obtain the merged \unionGAN.

\subsection{First stage: Model rooting}
\label{subsec:rooting}
Our goal is to merge the GAN models while maintaining as much information and generative performance as possible from the original models. In order to do so we need to somehow combine the weights of these models.

One way to combine several neural networks is by performing some arithmetic operations on their parameters.
For example, Exponential Moving Average (EMA) is a technique for averaging the model weights during the training process in order to merge several versions of the same model (from different checkpoints during the training process). EMA may be used for both discriminative \cite{tarvainen2017mean} and generative tasks \cite{karras2017progressive, stylegan1, stylegan2}.

In order to average the weights of several models, the weights must have the same dimensions.
However, this condition is not sufficient for achieving meaningful results.
For example, Figure \ref{fig:weights_averaging} (left) demonstrates that if we simply average the weights of two generators with the same architecture, which were trained on two different datasets, the resulting images have no apparent semantic structure.

\input{figures/weights_averaging/weights_averaging}

A key feature in the EMA case is that the averaging is performed on the same model from different training stages. Thus, we can say that the averaging is done on models that share the same \emph{common ancestor} model, and we hypothesize that this property is key to the success of the merging procedure.

Thus, given $N$ source GANs, $\{\gan{i} = (G_i, D_i)\}_{i=1}^{N}$, we (a) convert them to the same architecture (if their original architectures differ), and (b) create a common ancestor for all the models.
To meet these conditions we propose the model rooting technique: we choose one of the models arbitrarily (see \Cref{sec:choosing_root_model} for details) to be our root model $\gan{r}$; next, for each $i \in [N] \setminus r$ we train a model that is initialized by $\GAN_r$ on the outputs of $\gan{i}$, with the implicit task of performing catastrophic forgetting \cite{kirkpatrick2017overcoming} of the source dataset $r$. We denote each one of the resulting models as \ganArrow{r}{i}.
Now, models \gan{r} and \ganArrow{r}{i} not only share the same architecture but also share a common ancestor. Hence, averaging their weights will yield more semantically meaningful results, as demonstrated in \Cref{fig:weights_averaging} (right).

In order to quantify the distance between two models \gan{A} and \gan{B} we can measure the %
$L_2$ distance between their weights, i.e.:
$d(A, B) = \frac{1}{L}\sum_i\lVert\theta_{A,i} - \theta_{B, i}\rVert_2$ where $\theta_{A,i}$ is the $i$ layer of model $A$, $\theta_{B,i}$ is the $i$ layer of model $B$, and $L$ is the number of layers.
\Cref{fig:weights_averaging} (right) implies that the weights of \gan{A} are more aligned with those of \ganArrow{A}{B} than with the weights of \gan{B}. In order to verify this quantitatively, we report the distances $d(A, B)$ and $d(A,A \rightarrow B)$ in Table~\ref{tab:l2_weights_distance}. Indeed, the rooted models \ganArrow{A}{B} are much closer to the root, despite being trained on other datasets until convergence. Note that semantically closer datasets (e.g., cats and dogs) yield a smaller distance.

To conclude, the model rooting step transfers all the models to the same architecture and aligns their weights such that they can be averaged. Next, inspired by EMA, we will show that the averaging of the models introduces an inductive-bias to the training procedure that yields better results.

\input{tables/l2_weights_distance}

\subsection{Second stage: Model merging}
\label{subsec:finetune}
We now have $N$ rooted models, averaging whose weights yields somewhat semantically meaningful results. However, images generated by the averaged models are typically somewhere in between all the training classes (\Cref{fig:weights_averaging}, rightmost column). We want the model to learn to reuse filters that are applicable for all datasets, and differentiate the class-specific filters. For that, we continue with an adversarial training of the averaged model using the original GAN models as the data sources.

Specifically, given the $N$ rooted models from the previous stage: \gan{r} and $\{\ganArrow{r}{i}\}_{i \in [N] \setminus r}$ we create an average model: $\gan{a} = (G_a, D_a)$, s.t. $\theta_a = {\frac{1}{N}} (\theta_r + \sum_{i \in [N] \setminus r} \theta_{r \rightarrow i})$ where $\theta_i$ are the parameters of model $i$.
We also experimented with more sophisticated weighted average initialization based on the diagonal of the Fisher information matrix \cite{kirkpatrick2017overcoming} but it did not improve the results over a simple averaging. We then fine-tune $\gan{a}$ using the outputs of the original $\gan{i}$ models
to obtain the desired \unionGAN. 

%% file: tables/data_vs_synth.tex
\begin{table}[t]
\begin{center}
\caption{Comparison between FID scores of models that were trained on real and generated images}
 \begin{tabular}{lP{1.5cm}P{1.5cm}}
 \toprule
  & \multicolumn{2}{c}{\textbf{Trained on}} \\
 \cmidrule(lr){2-3} 
 \textbf{Dataset} & {real} & {generated} \\
 \midrule
 FFHQ & \bfseries \phantom{0}5.58 & \phantom{0}8.84 \\
 LSUN cat & \bfseries 17.37 & 21.78 \\
 LSUN dog & \bfseries 20.48 & 24.31 \\
 LSUN car & \bfseries \phantom{0}7.12 & 12.79 \\
 \bottomrule
\end{tabular}
\label{tab:real_vs_synth}
\end{center}
\end{table}

%% file: figures/weights_averaging/weights_averaging.tex
\begin{figure}[t]
    \centering
    \setlength{\tabcolsep}{0.5pt}
    \renewcommand{\arraystretch}{0.5}
    \newlength{\ww}
    \setlength{\ww}{0.15\textwidth}
  
    \begin{tabular}{ccc@{\hskip 0.6cm}ccc}
        \includegraphics[width=\ww,frame]{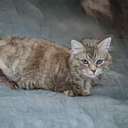} &
        \includegraphics[width=\ww,frame]{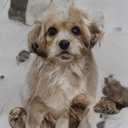} &
        \includegraphics[width=\ww,frame]{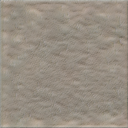} &
        \includegraphics[width=\ww,frame]{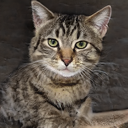} &
        \includegraphics[width=\ww,frame]{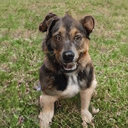} &
        \includegraphics[width=\ww,frame]{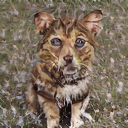} \\

        \includegraphics[width=\ww,frame]{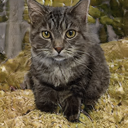} &
        \includegraphics[width=\ww,frame]{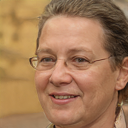} &
        \includegraphics[width=\ww,frame]{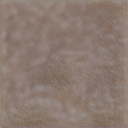} & \includegraphics[width=\ww,frame]{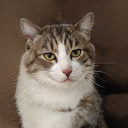} &
        \includegraphics[width=\ww,frame]{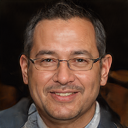} &
        \includegraphics[width=\ww,frame]{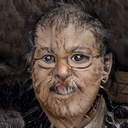} \\  
        
        \includegraphics[width=\ww,frame]{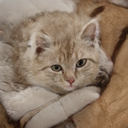} &
        \includegraphics[width=\ww,frame]{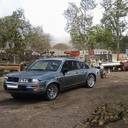} &
        \includegraphics[width=\ww,frame]{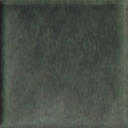} &
        \includegraphics[width=\ww,frame]{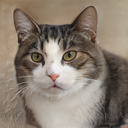} &
        \includegraphics[width=\ww,frame]{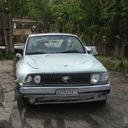} &
        \includegraphics[width=\ww,frame]{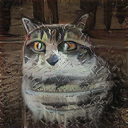} \\  
        
        \gan{i} & \gan{j} & Average & \gan{i} & \ganArrow{i}{j} & Average \\
        
    \end{tabular}
    
    \caption{\textbf{Left:} Averaging of two models $\gan{i}$ and $\gan{j} $ with the same architecture, but without a common root. 
    \textit{Average} is a model in which each weight is the arithmetic mean of the corresponding weights in the original two models.
    Each row corresponds to the same input $z$.
    Note that the resulting images exhibit no obvious semantic structure.
    \textbf{Right:} Averaging of two models which have a common root model. The resulting networks (before any fine-tuning) produce images which are a semantically meaningful mix of the two object categories.}
    \label{fig:weights_averaging}
\end{figure}

%% file: tables/l2_weights_distance.tex
\begin{table}[t]
\begin{center}
\caption{Rooted vs. not-rooted distance. The distance between the weights of the rooted model is much closer than that of the not-rooted model}
 \begin{tabular}{lcc}
 \toprule
Merged Datasets & $d(A, B)$ & $d(A,A \rightarrow B)$ \\
 \midrule
 cat + dog & 418.55 & \textbf{232.21} \\
 FFHQ + cat & 433.87 & \textbf{252.34} \\
 cat + car & 454.23 & \textbf{264.31} \\
 \bottomrule
\end{tabular}
\label{tab:l2_weights_distance}
\end{center}
\end{table}

%% file: sections/results.tex
\section{Results}
\label{sec:results}

Our main evaluation metric is the commonly-used Fr\'echet Inception Distance (FID) \cite{fid} which measures the Fr\'echet distance in the embedding space of the inception network between the real images and the generated images. The embedded data is assumed to follow a multivariate normal distribution, which is estimated by computing their mean
and covariance. We measure quality by computing FID
between 50k generated images and all of the available training images, as recommended by Heusel \etal~\cite{fid}.

All the FID scores that are reported in this paper were computed against the original training data. 
Note that this is for evaluation purposes only, and our models did not have access to the original data during training.

We evaluate our model on several representative cases using LSUN \cite{lsun} and FFHQ \cite{stylegan1} datasets. We specifically chose to compare between domains that are semantically close (cats and dogs), as well as domains that are more semantically distant (cats and cars). In addition, we compare aligned and unaligned datasets:
\begin{itemize}
    \itemsep0em 
    \item \textbf{Aligned and unaligned images:} we used LSUN cat dataset which contains images of cats in different poses and sizes, and FFHQ dataset which contains images of human faces that are strictly aligned. The FID between these two datasets is 196.59.
    \item \textbf{Unaligned imaged from related classes:} we used LSUN cat and LSUN dog classes. The FID between the two datasets is 72.2.
    \item \textbf{Unaligned imaged from unrelated classes:} we used LSUN cat and LSUN car classes. The FID between the two datasets is 161.62.
\end{itemize}
The FID distances reported above provide an indication of the semantic proximity between each pair of datasets. Not surprisingly, cat images are semantically closer to dog images than to images of humans (FFHQ) or of cars.

We compare our method against the following methods: training from scratch (\Cref{scratch}), TransferGAN (\Cref{sec:transfer_gan}), Elastic Weight Consolidation (\Cref{sec:ewc_baseline}) and the recently proposed Freeze Discriminator method \cite{mo2020freeze} which aims to improve transfer learning in GANs by freezing the highest-resolution layers of the discriminator during transfer.

In \Cref{tab:full_fid_results} we calculate the FID score between the union of all classes and the union of 50K samples of each class of the generated images. Our method outperforms other methods on all the datasets we experimented with.

\input{tables/fid_global}

In addition, we evaluated the FID score on each class separately in order to measure the effect of each method on each class. As can be seen in \Cref{tab:per_class_fid_comparison}, when the classes are semantically close, such as in the case of cat + dog, our method achieves better results than all the baselines. When the classes are semantically distant, such as in the case of cat + FFHQ or cat + car, we can see that EWC achieves better results on the class with respect to which it minimizes its weights distances, but this comes at the expense of the other class. This is the reason for the better overall performance of our method, reported in \Cref{tab:full_fid_results}.

\input{tables/per_class_fid_comparison}

As mentioned at the outset, the premise of this work is that the original training data is not available (which is the case with many real-world models). If the original training data is available to the merging process, the best merging results may unsurprisingly be achieved by simply training the new class-conditioned model on the union of the original training datasets. Results achieved in this manner are an upper bound for the results that can be achieved without access to the training data.
\Cref{tab:full_fid_results} and \Cref{tab:per_class_fid_comparison} show the gap between our merging approach (without training data) and the aforementioned upper bound.

\subsection{Choosing the root model}
\label{sec:choosing_root_model}
At the first stage of our approach we arbitrarily choose one of the models to serve as the root model. 
This raises the question of whether the choice of the root model matters. \Cref{tab:different_root_total_fid} shows that our method outperforms the baselines regardless of the model that is chosen as the root model.
On the other hand, it does not mean that the choice of the root model is insignificant for the overall performance of the final merged model. As we can see from \Cref{tab:different_root_total_fid}, when merging LSUN cat and LSUN dog models, the better overall result is achieved when LSUN dog is chosen as the root, while when merging the LSUN cat and FFHQ models, the better result is achieved by choosing LSUN cat to be the root model.

We hypothesized that a better candidate for the root model would be the generator that is more diverse, i.e., whose generated images are semantically far from each other. To test our hypothesis we calculated the diversity by measuring pairwise LPIPS scores between the generated images of each model. However, we found that it is not always the case that the more diverse generator is the better root model.

\input{tables/different_root_total_fid}

\subsection{Applications}
The output of our model is a single conditional GAN with a common latent space for all the classes. Hence, the merged model supports a variety of GAN applications from the literature. To name a few:

\textbf{Latent space interpolation} is used for demonstrating the smoothness of the latent space of a GAN. It can also be used for creating smooth transition sequences between objects of different classes.
In \Cref{fig:latent_interpolation} we demonstrate a transition between a cat and a dog by interpolating between their two $w$ latent vectors in the merged model from two different classes.

\input{figures/latent_interpolation/latent_interpolation}

\textbf{Style mixing}, introduced by Karras \etal~\cite{stylegan1}, is the ability the mix between generated images on different semantic levels (e.g., gender, hairstyle, pose, etc.) by feeding a different latent vector $w$ to different generation layers. Given a shared latent space for different classes enables us to use the style mixing mechanism to mix attributes from images belonging to these different classes, e.g., change the pose of a cat to that of a dog, while retaining the appearance of the cat. A few such examples are shown in \Cref{fig:style_mixing}. Note 
how both the pose and the general shape are taken from the source class
(e.g., the shape of the ears in column 1 is taken from the dog images, rather than from the cat images).

\input{figures/style_mixing/style_mixing}

\textbf{Semantic editing} is the ability to perform image editing operations on images by manipulating their latent space \cite{harkonen2020ganspace, shen2020interpreting, viazovetskyi2020stylegan2}. One advantage of our framework is the ability to edit images from different domains using the same latent direction because of the shared latent space. For example, given a model that merges FFHQ and LSUN cat generators,
we can leverage an off-the-shelf pose classifier, which is available for humans but not for cats, in order to classify poses as ``positive'' (pose from left to right) or ``negative'' (pose from right to left).
Applying this classifier only to images of humans generated by the merged model, we obtain a direction in the shared latent space that corresponds to a pose change, as the hyperplane normal of a linear SVM trained on the latent vectors of the human faces. \Cref{fig:semantic_editing} demonstrates that the same latent direction (that was calculated on humans only) can be applied for both humans and cats. So, using our model merging solution we can leverage off-the-shelf classifiers on one class to operate on all of the classes.

\input{figures/semantic_editing/semantic_editing}

%% file: tables/fid_global.tex
\begin{table}[t]
\begin{center}
\caption{Comparison of FID score w.r.t. the union of all the datasets, for several dataset combinations. Cat, dog, and car datasets are taken from LSUN \cite{lsun}}
 \begin{adjustbox}{width=0.8\columnwidth}
 \begin{tabular}{lccccc} \toprule
 Datasets & 
 \begin{tabular}{c} FFHQ \\ cat \\ \end{tabular} &
 \begin{tabular}{c} cat \\ dog \\ \end{tabular} &
 \begin{tabular}{c} cat \\ car \\ \end{tabular} &
 \begin{tabular}{c} FFHQ \\ cat \\ dog \end{tabular} &
 \begin{tabular}{c} FFHQ \\ cat \\ dog \\ car \\ \end{tabular} \\
 [0.5ex] 
 \midrule
 From scratch & 19.61 & 27.58 & 20.52 & 23.22 & 24.88 \\ 
 TransferGAN \cite{transferring_gans} & 18.63 & 22.17 & 17.77 & 20.64 & 19.34 \\ 
 EWC \cite{kirkpatrick2017overcoming} & 19.45 & 22.17 & 17.65 & 19.47 & 19.14 \\ 
 Freeze-D \cite{mo2020freeze} & 18.17 & 21.92 & 17.52 & 19.71 & 19.41 \\ 
 \midrule
 Our & \textbf{16.44} & \textbf{20.77} & \textbf{16.85} & \textbf{18.98} & \textbf{18.44} \\
 \midrule
 Upper bound (real data training) & 11.86 & 17.68 & 14.28 & 15.93 & 16.45 \\ 
 \bottomrule
\end{tabular}
\end{adjustbox}
\label{tab:full_fid_results}
\end{center}
\end{table}

%% file: tables/per_class_fid_comparison.tex
\begin{table}[t]
\begin{center}
 \caption{FID scores per-class over different dataset combinations}
 \begin{adjustbox}{width=1.0\columnwidth}
 \begin{tabular}{lcccccc}
 \toprule
 & \multicolumn{2}{c}{\textbf{LSUN cat+dog}} & \multicolumn{2}{c}{\textbf{LSUN cat+car}}  & \multicolumn{2}{c}{\textbf{LSUN cat+FFHQ}} \\
\cmidrule(lr){2-3} \cmidrule(lr){4-5} \cmidrule(lr){6-7}
 \textbf{Dataset} & cat & dog & cat & car & FFHQ & cat \\
 [0.5ex] 
 \midrule
 Scratch & 30.37 & 33.21 & 32.21 & 14.43 & 13.35 & 31.64 \\ 
 TransferGAN \cite{transferring_gans} & 23.32 & 28.84 & 30.06 & 11.49 & 11.16 & 32.08 \\ 
 EWC \cite{kirkpatrick2017overcoming} & 23.04 & 30.11 & 30.65 & \textbf{10.54} & \textbf{9.85} & 35.36 \\
 Freeze-D \cite{mo2020freeze}         & 23.36 & 28.40 & 29.78 & 11.44 & 10.64 & 31.57 \\ 
\midrule
Our & \textbf{22.08} & \textbf{26.52} & \textbf{27.78} & 11.59 & 10.60 & \textbf{27.82} \\
\midrule
Upper bound (real data training) & 16.49 & 24.75 & 23.23 & 9.5 & 8.49 & 19.19 \\ 
\bottomrule
\end{tabular}
\end{adjustbox}
\label{tab:per_class_fid_comparison}
\end{center}
\end{table}

%% file: tables/different_root_total_fid.tex
\begin{table}[t]
\begin{center}
\caption{Our method outperforms the baselines, in terms of FID score, regardless of the model that is chosen as the root model.}
 \begin{tabular}{lcccccc}
 \toprule
 & \multicolumn{2}{c}{\textbf{LSUN cat + LSUN dog}} & \multicolumn{2}{c}{\textbf{LSUN cat + FFHQ}} \\
\cmidrule(lr){2-3} \cmidrule(lr){4-5} 
 \textbf{Root model} & LSUN cat & LSUN dog & FFHQ & LSUN cat \\
 [0.5ex] 
 \midrule
 Scratch & 27.58 & 27.58 & 19.61 & 19.61 \\ 
 TransferGAN \cite{transferring_gans} & 22.17 & 21.11 & 18.63 & 16.40 \\ 
 EWC \cite{kirkpatrick2017overcoming} & 22.17 & 25.16 & 19.45 & 16.52 \\
 Freeze-D \cite{mo2020freeze}         & 21.92 & 25.03 & 18.17 & 16.87 \\ 
\midrule
Our & \textbf{20.77} & \textbf{20.03} & \textbf{16.44} & \textbf{15.60} \\
\bottomrule
\end{tabular}
\label{tab:different_root_total_fid}
\end{center}
\end{table}

%% file: figures/latent_interpolation/latent_interpolation.tex
\begin{figure}[t]
    \centering
    \setlength{\tabcolsep}{0.5pt}
    \renewcommand{\arraystretch}{0.5}
    \setlength{\ww}{0.135\textwidth}
  
    \begin{tabular}{ccccccc}
        \includegraphics[width=\ww,frame]{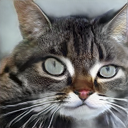} &
        \includegraphics[width=\ww,frame]{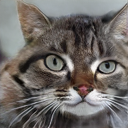} &
        \includegraphics[width=\ww,frame]{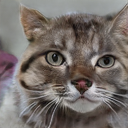} &
        \includegraphics[width=\ww,frame]{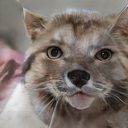} &
        \includegraphics[width=\ww,frame]{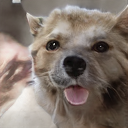} &
        \includegraphics[width=\ww,frame]{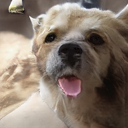} &
        \includegraphics[width=\ww,frame]{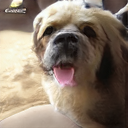}
    \end{tabular}
    
    \caption{Interpolation in the merged model's latent space of between images of different classes.}
    \label{fig:latent_interpolation}
\end{figure}

%% file: figures/style_mixing/style_mixing.tex
\begin{figure}[t]
    \centering
    \setlength{\tabcolsep}{0.5pt}
    \renewcommand{\arraystretch}{0.5}
    \setlength{\ww}{0.135\textwidth}

    \begin{tabular}{cccccc}
        {\scriptsize Appearance} \rotatebox{90}{\scriptsize\phantom{AAA}Pose} &
        \includegraphics[width=\ww,frame]{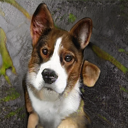} &
        \includegraphics[width=\ww,frame]{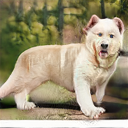} &
        \includegraphics[width=\ww,frame]{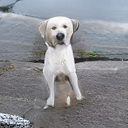} &
        \includegraphics[width=\ww,frame]{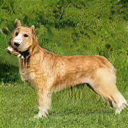} &
        \includegraphics[width=\ww,frame]{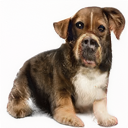} \\
        
        \includegraphics[width=\ww,frame]{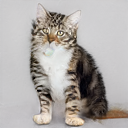} &
        \includegraphics[width=\ww,frame]{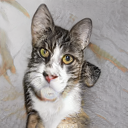} &
        \includegraphics[width=\ww,frame]{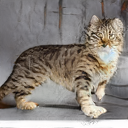} &
        \includegraphics[width=\ww,frame]{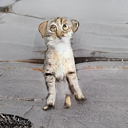} &
        \includegraphics[width=\ww,frame]{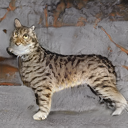} &
        \includegraphics[width=\ww,frame]{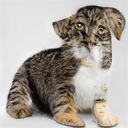} \\
        
        \includegraphics[width=\ww,frame]{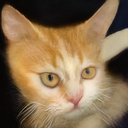} &
        \includegraphics[width=\ww,frame]{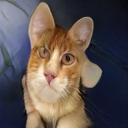} &
        \includegraphics[width=\ww,frame]{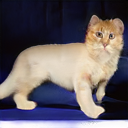} &
        \includegraphics[width=\ww,frame]{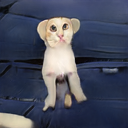} &
        \includegraphics[width=\ww,frame]{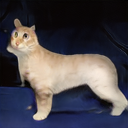} &
        \includegraphics[width=\ww,frame]{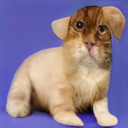} \\
        
        \includegraphics[width=\ww,frame]{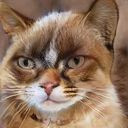} &
        \includegraphics[width=\ww,frame]{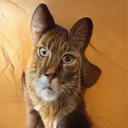} &
        \includegraphics[width=\ww,frame]{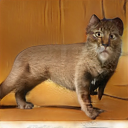} &
        \includegraphics[width=\ww,frame]{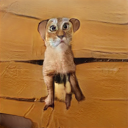} &
        \includegraphics[width=\ww,frame]{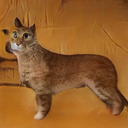} &
        \includegraphics[width=\ww,frame]{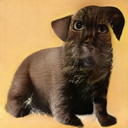} \\
        
        \includegraphics[width=\ww,frame]{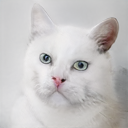} &
        \includegraphics[width=\ww,frame]{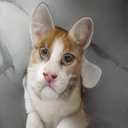} &
        \includegraphics[width=\ww,frame]{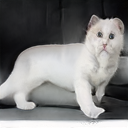} &
        \includegraphics[width=\ww,frame]{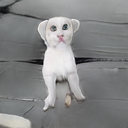} &
        \includegraphics[width=\ww,frame]{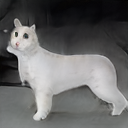} &
        \includegraphics[width=\ww,frame]{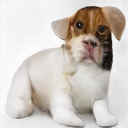} \\
        
        \includegraphics[width=\ww,frame]{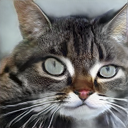} &
        \includegraphics[width=\ww,frame]{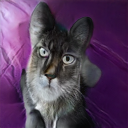} &
        \includegraphics[width=\ww,frame]{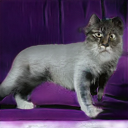} &
        \includegraphics[width=\ww,frame]{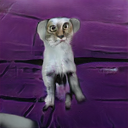} &
        \includegraphics[width=\ww,frame]{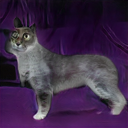} &
        \includegraphics[width=\ww,frame]{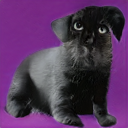} \\
    \end{tabular}
    
    \caption{Style mixing between images of two different domains: taking the pose and shape from the dog image and the appearance from the cat image.}
    \label{fig:style_mixing}
\end{figure}

%% file: figures/semantic_editing/semantic_editing.tex
\begin{figure}[t]
    \centering
    \setlength{\tabcolsep}{0.5pt}
    \renewcommand{\arraystretch}{0.5}
    \setlength{\ww}{0.153\textwidth}
  
    \begin{tabular}{ccccc}
        \includegraphics[width=\ww,frame]{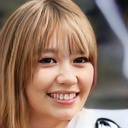} &
        \includegraphics[width=\ww,frame]{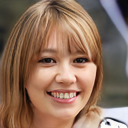} &
        \includegraphics[width=\ww,frame]{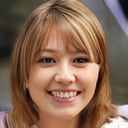} &
        \includegraphics[width=\ww,frame]{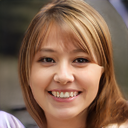} &
        \includegraphics[width=\ww,frame]{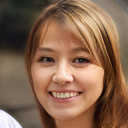} \\
        
        \includegraphics[width=\ww,frame]{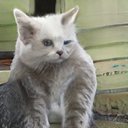} &
        \includegraphics[width=\ww,frame]{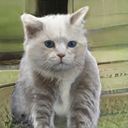} &
        \includegraphics[width=\ww,frame]{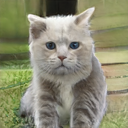} &
        \includegraphics[width=\ww,frame]{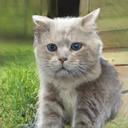} &
        \includegraphics[width=\ww,frame]{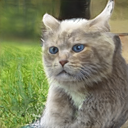} \\
        
        (-) Pose &  & Neutral &  & (+) Pose
    \end{tabular}
    
    \caption{We determine the pose direction in the latent space of the merged model of FFHQ and LSUN cat using images of the FFHQ class only. Applying this direction to images from both classes reveals that the semantics stay the same.}
    \label{fig:semantic_editing}
\end{figure}

%% file: sections/limitations.tex
\section{Limitations and Future Work}
\label{sec:limitations}

Due to our self-imposed constraint on model size, we have found that our solution (and the baselines) are sensitive to the number of source models and their properties: merging more models or merging models with semantically distant distributions produces higher FID scores, as may be seen in \Cref{tab:cat_fid_over_datasets}. For example, merging LSUN cat and LSUN dog produced better FID scores on the cat dataset on all the baselines in comparison to merging LSUN cat and LSUN car. We conjecture that this happens because more filters can be reused among the semantically similar datasets. Additionally, we can see that merging four of the datasets produces the worst result on our method.
Notice that merging FFHQ + cat + dog produced a better result than cat + car and FFHQ + cat because of the semantical closeness of cat and dog.

Yet another disadvantage of our method is that the training time is longer due to the two-stage approach. The baselines converge faster to their local minima, but result in a higher FID score than our method.

\input{tables/cat_FID_over_datasets}

Future work can be to relax the capacity constraint and allow a minimal capacity and run-time increase to enable merging more models or models from semantically distant distributions with better results in terms of FID score.

%% file: tables/cat_FID_over_datasets.tex
\begin{table}[t]
\begin{center}
\caption{Comparison of FID scores of cat class only, when merging LSUN cat with different datasets. Semantically closer datasets (e.g., cats and dogs) lead to better scores compared to far datasets (e.g., cats and cars). Merging 4 datasets produces the worst result}
 \begin{tabular}{lccccc}
 \toprule
 Datasets & 
 \begin{tabular}{c} cat \\ dog \\ \end{tabular} &
 \begin{tabular}{c} cat \\ car \\ \end{tabular} &
 \begin{tabular}{c} FFHQ \\ cat \\ \end{tabular} &
 \begin{tabular}{c} FFHQ \\ cat \\ dog \end{tabular} &
 \begin{tabular}{c} FFHQ \\ cat \\ dog \\ car \\ \end{tabular} \\
 [0.5ex] 
 \midrule
 From scratch & 30.37 & 32.21 & 31.64 & 34.75 & 45.02 \\ 
 TransferGAN \cite{transferring_gans} & 23.32 & 30.06 & 32.08 & 29.06 & 30.50 \\ 
 EWC \cite{kirkpatrick2017overcoming} & 23.04 & 30.65 & 35.36 & 25.70 & 27.98 \\ 
 Freeze-D \cite{mo2020freeze} & 23.36 & 29.78 & 31.57 & 28.68 & 30.95 \\ 
 \midrule
 Our & 22.08 & 27.78 & 27.82 & 26.80 & 30.17 \\ 
 \bottomrule
\end{tabular}
\label{tab:cat_fid_over_datasets}
\end{center}
\end{table}

%% file: sections/social_implications.tex
\section{Broader Impact}
\label{sec:social_implications}

One major barrier when developing a machine learning model is the lack of training data. Many small organizations and individuals find it hard to compete with larger entities due to the lack of training data. This is especially true in fields where curating and annotating the training data is time-consuming and expensive (e.g., medical data). It was shown that GANs can be used in order to augment and anonymize sensitive training data \cite{shin2018medical, yoon2020anonymization}. Our method can be used 
to alleviate the problem of scarce training data, by allowing entities with small budgets to use  
pre-trained GANs in two ways: use them as training data, and reuse some of the knowledge incorporated in their weights.

On the other hand, our method may amplify the copyright issues that arise when training a model on synthetic data. The legal implications of training a model using another model that was trained on a private or copyrighted dataset are currently unclear. We would like to encourage the research community to work with governments and legal scholars to establish new laws and regulations in lockstep with the rapid advancement of synthetic media.

%% file: sections/conclusions.tex
\section{Conclusions}
\label{sec:conclutions}

In this paper, we introduced the problem of merging several generative adversarial networks without having access to the training data. We adapted current methods for transfer-learning and continual-learning
and set them as our baselines. We then introduced our novel two-stage solution to the GAN mixing problem: model rooting and model merging. Later, we compared our method to the baselines and demonstrated its superiority on various datasets. Finally, we presented some applications of our model merging technique.

%% file: appendices/implementation_details.tex
\section{Implementation Details}
We evaluated our technique and the baselines using the StyleGAN2 architecture \cite{stylegan2}. We kept most of the details unchanged, including network architecture, weight demodulation, regularization, exponential moving average of generator weights, R1 regularization~\cite{mescheder2018training}, mini-batch size of 32 images, and using the Adam optimizer~\cite{kingma2015adam} with $\beta_1 = 0, \beta_2=0.99$ and $\epsilon=10^{-8}$.

In order to introduce conditioning to the unconditioned StyleGAN2 architecture we add the following components to the generator and the discriminator:
\begin{itemize}
    \item \textbf{Generator conditioning.} We add a class embedding layer to the mapping network of the generator, s.t. the input to the generator is noise vector $z$ and one-hot class $c$. The embedded condition is concatenated to the input $z$. The first fully-connected layer of the mapping network is modified to support this new size.
    \item \textbf{Discriminator conditioning.} We add a mapping network to the discriminator
    that gets as an input only a class one-hot vector $c$ (with no noise vector $z$) and calculates a $w$ vector. We then incorporate this $w$ vector to the final discriminator prediction by a projection \cite{miyato2018cgans}.
\end{itemize}

If our input models are unconditioned or conditioned with an insufficient number of classes, we can easily introduce/extend the class embedding layer to the input models to the desired size by adding more rows to the embedding matrix and initialize it randomly.

It is important to notice that we do not rely on any of StyleGAN's features in our solution (or in the baselines), so our solution is agnostic to the input GAN architecture.

\subsection{Hyperparameters and training configurations}
We used the same hyperparameter configurations as in the PyTorch \cite{paszke2019pytorch} implementation of StyleGAN2-ada \cite{karras2020training}, while we did not use the adaptive augmentation capabilities. We used a fixed mapping depth of 8 layers during all our experiments. The hyperparameters were chosen by a random search and are available at the source code.

We used a single NVIDIA RTX 2080 GPU per experiment.
We incorporated mixed-precision training~\cite{micikevicius2017mixed} in order to speed up the training process.

We trained all our experiments until convergence, which takes about 5M training steps because we start from pre-trained models. Each stage of our two-stage approach (model rooting and merging) takes about 2 days on NVIDIA RTX 2080, thus the total training time is about 4 days.

%% file: appendices/additional_experiments.tex
\section{Additional experiments}
In addition to the experiments reported in the main paper we also compared our method on other datasets.
Furthermore, we experimented with mixing models of different architectures and mixing models of different quality.

\subsection{Additional datasets}
We compared our method using additional classes from the LSUN dataset. As can be seen in \Cref{tab:additional_full_fid_results}, our method outperforms the baselines in all of our experiments. In addition, we tested the effect of using multiple source datasets, as reported in \Cref{tab:fid_global_multiple_datasets}. As we can see, our method outperforms the baselines even when mixing seven different models.

\input{tables/fid_global_additional_datasets.tex}
\input{tables/fid_global_multiple_datasets.tex}

\subsection{Mixing models with different architectures}
\label{sec:different_architectures}
For most of our experiments we use the StyleGAN2 framework.
However, our method can be used to merge models with different architectures. In the first stage (model rooting) after we choose the root model, the remaining models serve only as data-generators, hence can be of any architecture.  In the second stage (model merging), all the rooted models that we create are, by design, of the same architecture as the root.

\input{tables/different_architectures}

We evaluated our solution (and the baselines) on merging models with different architectures: a StyleGAN2 model trained on LSUN cat and a custom made model that was trained on LSUN dog. The custom made model was created by removing the mapping network from the StyleGAN architecture and replace it with a simple linear embedding layer. Each of the models was chosen as root, thus in one case the merged model is a StyleGAN2, and in the other case, the merged model is a custom one.
As shown in \Cref{tab:different_architectures}, our mixing approach outperforms the baselines, in terms of FID score, regardless of the architecture of the root model. Again, it does not mean that the root model is meaningless: choosing the StyleGAN2 architecture for the merged model produces superior results, compared to merging that uses the custom architecture.

We have noticed that both of the source models have comparable FID scores, which leads us to the next question: what happens if we mix models of different FID scores.

\subsection{Mixing models of different quality}
\label{sec:different_dataset_sizes}
In order to isolate and identify changes that result in consistent improvements across our various experiments, we mainly focus on comparing models of the same quality: models of the same capacity that were trained on roughly the same dataset size. This raises the question of whether our method is beneficial in the cases where the models are of different quality.

To test under such conditions, we trained a StyleGAN2 model on a reduced version of LSUN dog with an order of magnitude fewer training samples: we evaluated the mixing between a model that was trained on 100K samples of LSUN cat and a model that was trained on 10K samples of LSUN dog.
\Cref{tab:different_dataset_sizes} demonstrates that in this scenario, our method still outperforms the baselines regardless of the choice of the root model. It is also important to notice that EWC performs significantly worse when the root model is the one that was trained on the smaller dataset, because the inductive bias towards the weights of this weaker model is a bad prior. The other baselines, as well as our method, are much less sensitive. Nevertheless, we can see that choosing the root model to be the model that was trained on the larger dataset yields better results.

\input{tables/different_dataset_sizes}

%% file: tables/fid_global_additional_datasets.tex
\begin{table}[t]
\begin{center}
\caption{Comparison of FID score w.r.t. the union of all the datasets, for several dataset combinations. Cat, horse, church and bedroom datasets are taken from LSUN \cite{lsun}}
 \begin{adjustbox}{width=1\columnwidth}
 \begin{tabular}{lccccccccc} \toprule
 Datasets & 
 \begin{tabular}{c} cat \\ horse \\ \end{tabular} &
 \begin{tabular}{c} cat \\ bedroom \\ \end{tabular} &
 \begin{tabular}{c} cat \\ church \\ \end{tabular} &
 \begin{tabular}{c} FFHQ \\ horse \end{tabular} &
 \begin{tabular}{c} FFHQ \\ bedroom \end{tabular} &
 \begin{tabular}{c} FFHQ \\ church \end{tabular} &
 \begin{tabular}{c} horse \\ church \end{tabular} &
 \begin{tabular}{c} horse \\ bedroom \end{tabular} &
 \begin{tabular}{c} church \\ bedroom \end{tabular}
 \\
 [0.5ex] 
 \midrule
 From scratch & 20.53 & 22.81 & 21.01 & 13.74 & 14.96 & 11.83 & 12.85 & 16.96 & 13.33 \\ 
 TransferGAN \cite{transferring_gans} & 16.73 & 18.7 & 17.22 & 14.4 & 13.88 & 11.19 & 12.27 & 15.22 & 11.84 \\ 
 EWC \cite{kirkpatrick2017overcoming} & 17.46 & 17.98 & 16.75 & 13.27 & 14.25 & 11.11 & 14.51 & 15.43 & 11.07 \\ 
 Freeze-D \cite{mo2020freeze} & 16.98 & 18.53 & 18.04 & 13.25 & 13.14 & 9.74 & 10.78 & 15.81 & 11.81 \\ 
 \midrule
 Our & \textbf{16.46} & \textbf{16.7} & \textbf{15.62} & \textbf{11.28} & \textbf{11.5} & \textbf{9.52} & \textbf{10.61} & \textbf{13.9} & \textbf{9.91} \\ 
 \midrule
 Upper bound (real data training) & 13.17 & 14.42 & 13.65 & 8.52 & 8.59 & 6.25 & 8.65 & 10.26 & 7.01 \\ 
 \bottomrule
\end{tabular}
\end{adjustbox}
\label{tab:additional_full_fid_results}
\end{center}
\end{table}

%% file: tables/fid_global_multiple_datasets.tex
\begin{table}[t]
\begin{center}
\caption{Comparison of FID score w.r.t. the union of all the datasets, for several dataset combinations. Cat, horse, church, car and bedroom datasets are taken from LSUN \cite{lsun}.}
 \begin{adjustbox}{width=1\columnwidth}
 \begin{tabular}{lcccccc} \toprule
 Datasets & 
 \begin{tabular}{c} FFHQ \\ cat \\ \end{tabular} &
 \begin{tabular}{c} FFHQ \\ cat \\ dog \end{tabular} &
 \begin{tabular}{c} FFHQ \\ cat \\ dog \\ car \\ \end{tabular} &
 \begin{tabular}{c} FFHQ \\ cat \\ dog \\ car \\ horse \\ \end{tabular} &
 \begin{tabular}{c} FFHQ \\ cat \\ dog \\ car \\ horse \\ bedroom \\ \end{tabular} &
 \begin{tabular}{c} FFHQ \\ cat \\ dog \\ car \\ horse \\ bedroom \\ church \\ \end{tabular}
 \\
 [0.5ex] 
 \midrule
 From scratch & 19.61 & 23.22 & 24.88 & 20.56 & 18.36 & 18.34 \\ 
 TransferGAN \cite{transferring_gans} & 18.63 & 20.64 & 19.34 & 18.4 & 17.49 & 17.29 \\ 
 EWC \cite{kirkpatrick2017overcoming} & 19.45 & 19.47 & 19.14 & 18.14 & 18.56 & 17.18 \\ 
 Freeze-D \cite{mo2020freeze} & 18.17 & 19.71 & 19.41 & 17.8 & 17.24 & 17.5 \\ 
 \midrule
 Our & \textbf{16.44} & \textbf{18.98} & \textbf{18.44} & \textbf{17.35} & \textbf{17.04} & \textbf{16.41} \\
 \midrule
 Upper bound (real data training) & 11.86 & 15.93 & 16.45 & 16.19 & 16.88 & 16.33 \\ 
 \bottomrule
\end{tabular}
\end{adjustbox}
\label{tab:fid_global_multiple_datasets}
\end{center}
\end{table}

%% file: tables/different_architectures.tex
\begin{table}[t]
\begin{center}
\caption{\textbf{Mixing models of different architectures.} Our method outperforms the baselines, in terms of FID score, regardless of the architecture of the chosen root model}
 \begin{tabular}{lcccccc}
 \toprule
 & \multicolumn{2}{c}{\textbf{LSUN cat + LSUN dog}} \\
\cmidrule(lr){2-3}
 \textbf{Root model} & LSUN cat & LSUN dog \\
& (StyleGAN) & (Custom) \\
 [0.5ex] 
 \midrule
 Scratch & 23.34 & 23.34 \\ 
 TransferGAN \cite{transferring_gans} &  19.76 & 22.39 \\ 
 EWC \cite{kirkpatrick2017overcoming} & 21.57 & 21.79 \\
 Freeze-D \cite{mo2020freeze}         & 19.70 & 21.44 \\ 
\midrule
Our & \textbf{19.42} & \textbf{21.28} \\
\bottomrule
\end{tabular}
\label{tab:different_architectures}
\end{center}
\end{table}

%% file: tables/different_dataset_sizes.tex
\begin{table}[t]
\begin{center}
\caption{\textbf{Mixing models of different dataset sizes.} Our method outperforms the baselines, in terms of FID score, regardless of the model that is chosen as a root model. In addition, we can see that EWC performs poorly when initialized with the weaker model}
 \begin{tabular}{lcccccc}
 \toprule
 & \multicolumn{2}{c}{\textbf{LSUN cat + LSUN dog}} \\
\cmidrule(lr){2-3}
 \textbf{Root model} & LSUN cat & LSUN dog \\
\textbf{\# Training samples}& (100K) & (10K) \\
 [0.5ex] 
 \midrule
 Scratch & 37.46 & 37.46 \\ 
 TransferGAN \cite{transferring_gans} &  32.52 & 34.93 \\ 
 EWC \cite{kirkpatrick2017overcoming} & 32.18 & 45.10 \\
 Freeze-D \cite{mo2020freeze}         & 32.03 & 35.30 \\ 
\midrule
Our & \textbf{31.71} & \textbf{32.59} \\
\bottomrule
\end{tabular}
\label{tab:different_dataset_sizes}
\end{center}
\end{table}

%% file: appendices/datasets.tex
\section{Datasets}
We used FFHQ \cite{stylegan1} and LSUN \cite{lsun} datasets for our experiments. We used the entire FFHQ dataset which contains 70K images that are automatically aligned and cropped.

Images in the FFHQ dataset are licensed under either Creative Commons BY 2.0, Creative Commons BY-NC 2.0, Public Domain Mark 1.0, Public Domain CC0 1.0, or U.S. Government Works license. All of these licenses allow free use, redistribution, and adaptation for non-commercial purposes.

The LSUN dataset contains around one million labeled images for each of 10 scene categories and 20 object
categories. We used only some of the categories in the dataset (cat, dog, and car) and used only 100K images per class (in order to keep the balance between the FFHQ and the LSUN classes).

We trained the models once and then used the output of the trained models for our experiments. The dataset was used during our experiments only for calculation of the FID metrics. Note that we did not change the behavior of the training process based on the FID score in any way, because we assume that our method should be applicable without any training data. The multivariate Gaussian statistics of the inception features may not be available during the training for the end-user, hence we cannot use it.

\subsection{FID calculations}
\label{sec:fid_calculations}
The results in the tables in the main paper
are calculated over images of size $64\times64$, for efficiency reasons. 
To make sure that the same trends hold for higher resolutions, we tested our method on images of sizes $128\times128$ and $256\times256$ on the LSUN cat and LSUN dog datasets and achieved similar results, as can be seen in
\Cref{tab:full_fid_results_128}.
Each stage of our two-stage approach (model rooting and merging) takes about 4 days on NVIDIA RTX 2080 for resolution $128\times128$, and about 7 days on NVIDIA A10 for resolution $256\times256$; thus, the total training time is about 8 days and 14 days, respectively.

\input{tables/fid_global_128}

%% file: tables/fid_global_128.tex
\begin{table}[t]
\begin{center}
\caption{FID comparison of merging LSUN cat + LSUN dog when training on a higher resolutions}
 \begin{tabular}{lc|c}
 \toprule
 Datasets resolution & 
 $128\times128$ &
 $256\times256$ \\
 \midrule
 From scratch & 32.58 & 28.61 \\ 
 TransferGAN \cite{transferring_gans} & 26.90 & 23.28 \\ 
 EWC \cite{kirkpatrick2017overcoming} & 28.26 & 27.18 \\ 
 Freeze-D \cite{mo2020freeze} & 29.00 & 23.65 \\ 
 \midrule
 Our & \textbf{25.12} & \textbf{22.45} \\ 
 \midrule
 Upper bound (real data training) & 18.82 & 18.88 \\ 
 \bottomrule
\end{tabular}
\label{tab:full_fid_results_128}
\end{center}
\end{table}

%% file: appendices/training.tex
\section{Training}
In \Cref{fig:cat_dog_total_fid_convergence_rate} we can see the convergence rate of the FID that is calculated on the union of the input datasets LSUN cat and LSUN dog (which are semantically close datasets) during the training process. As we can see, our solution converges more quickly and to a lower FID than the baselines.

\input{figures/cat_dog_convergence/full}

In \Cref{fig:cat_dog_per_class_fid_convergence_rate} we show the FID score that is calculated per class. As we can see, TransferGAN is suffering from catastrophic forgetting on the cat class (left) that is somewhat mitigated by the EWC but it comes at the expense of increasing the FID score of the dog class (right). In contrast, our method starts from a higher FID score on the cat dataset than TrasferGAN/EWC/Freeze-D methods (because they started from the pretrained cat model which achieves better results), but later on, our method achieves a better result.

\input{figures/cat_dog_convergence/per_class}

In \Cref{fig:cat_ffhq_total_fid_convergence_rate} we can see the convergence rate of the total FID score when merging two semantically distant datasets: FFHQ and LSUN cat. We can see that our method converges more quickly and to a lower FID than the baselines. As we can see in \Cref{fig:cat_ffhq_per_class_fid_convergence_rate} again, EWC mitigates the catastrophic forgetting of TransferGAN on the FFHQ class and even achieves a better result on this class than our method. But it achieves the worst result on the second class (even worse than the \naive{} from scratch approach). So all-in-all it is outperformed by our method as can be seen in \Cref{fig:cat_ffhq_total_fid_convergence_rate}.

\input{figures/cat_ffhq_convergence/full}
\input{figures/cat_ffhq_convergence/per_class}

%% file: figures/cat_dog_convergence/full.tex
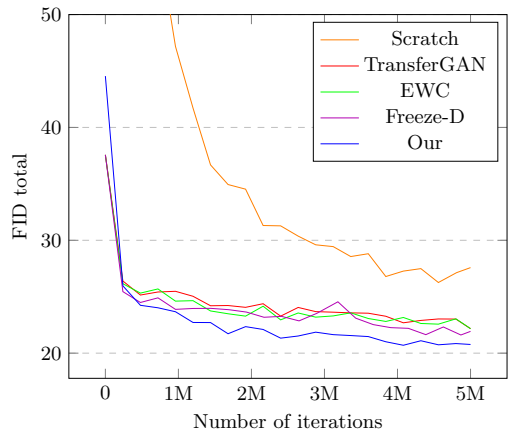
\begin{figure}[t]
\centering
\begin{tikzpicture}[scale=0.85]
\begin{axis}[
    xlabel={Number of iterations},
    ylabel={FID total},
    xticklabels={X, 0, 1M, 2M, 3M, 4M, 5M},
    ymax=50,
    ymajorgrids=true,
    grid style=dashed,
]

\addplot[color=orange]
coordinates {
(0.0,312.47)(240.0,98.94)(480.0,71.08)(720.0,56.77)(960.0,47.18)(1200.0,41.72)(1440.0,36.68)(1680.0,34.94)(1920.0,34.53)(2160.0,31.31)(2400.0,31.28)(2640.0,30.37)(2880.0,29.60)(3120.0,29.44)(3360.0,28.56)(3600.0,28.81)(3840.0,26.79)(4080.0,27.27)(4320.0,27.49)(4560.0,26.26)(4808.0,27.13)(5000.0,27.58)
};

\addplot[color=red]
coordinates {
(0.0,37.57)(240.0,26.39)(480.0,25.15)(720.0,25.42)(960.0,25.48)(1200.0,25.04)(1440.0,24.20)(1680.0,24.22)(1920.0,24.06)(2160.0,24.38)(2400.0,23.28)(2640.0,24.05)(2880.0,23.68)(3120.0,23.63)(3360.0,23.57)(3600.0,23.54)(3840.0,23.27)(4080.0,22.69)(4320.0,22.90)(4560.0,23.03)(4800.0,23.02)(5000.0,22.17)
};

\addplot[color=green]
coordinates {
(0.0,37.57)(240.0,26.15)(480.0,25.32)(720.0,25.69)(960.0,24.61)(1200.0,24.65)(1440.0,23.74)(1680.0,23.49)(1920.0,23.28)(2160.0,24.16)(2400.0,22.95)(2640.0,23.56)(2880.0,23.19)(3120.0,23.30)(3360.0,23.57)(3600.0,23.07)(3840.0,22.81)(4080.0,23.16)(4320.0,22.63)(4560.0,22.57)(4800.0,23.04)(5000.0,22.17)
};

\addplot[color=cyan]
coordinates {
(0.0,37.57)(240.0,25.45)(480.0,24.49)(720.0,24.90)(960.0,23.88)(1200.0,23.96)(1444.0,23.97)(1692.0,23.85)(1932.0,23.64)(2172.0,23.18)(2412.0,23.26)(2652.0,22.86)(2896.0,23.53)(3184.0,24.55)(3428.0,23.10)(3668.0,22.53)(3908.0,22.26)(4148.0,22.20)(4388.0,21.64)(4628.0,22.32)(4868.0,21.61)(5000.0,21.93)
};

\addplot[color=blue]
coordinates {
(0.0,44.55)(240.0,25.96)(480.0,24.24)(720.0,24.02)(960.0,23.67)(1200.0,22.72)(1440.0,22.71)(1680.0,21.72)(1920.0,22.35)(2160.0,22.10)(2400.0,21.33)(2640.0,21.52)(2880.0,21.86)(3120.0,21.64)(3360.0,21.56)(3600.0,21.47)(3840.0,21.01)(4080.0,20.70)(4320.0,21.10)(4560.0,20.74)(4800.0,20.85)(5000.0,20.77)
};

\legend{Scratch, TransferGAN, EWC, Freeze-D, Our}
    
\end{axis}
\end{tikzpicture}
\caption{Convergence rate of the total FID score during the training on LSUN cat + LSUN dog. We can see that our solution achieves the lowest (best) FID score.}
\label{fig:cat_dog_total_fid_convergence_rate}
\end{figure}

%% file: figures/cat_dog_convergence/per_class.tex
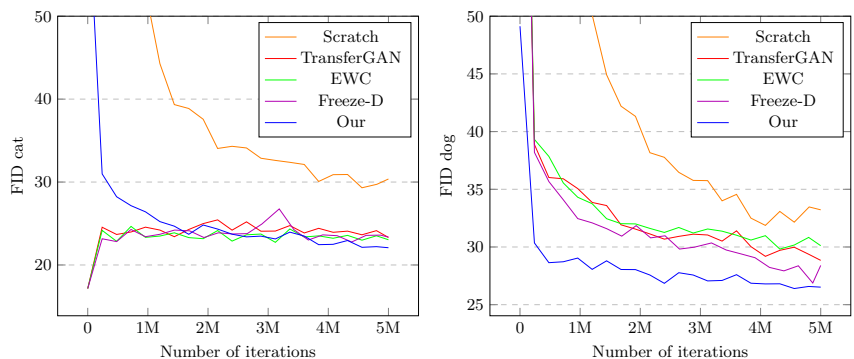
\begin{figure}[t]
    \centering
    \begin{tabular}{cc}
        \input{figures/cat_dog_convergence/cat}
        & 
        \input{figures/cat_dog_convergence/dog}
    \end{tabular}
    \caption{The FID score that is calculated on the cat class (left) and on the dog class (right). As we can see, the TransferGAN suffers from catastophic forgetting of the cat class (left) that is somewhat mitigated by the EWC but it comes at the expense of increasing the FID score of the dog class (right).}
    \label{fig:cat_dog_per_class_fid_convergence_rate}
\end{figure}

%% file: figures/cat_dog_convergence/cat.tex
\begin{tikzpicture}[scale=0.7]
\begin{axis}[
    xlabel={Number of iterations},
    ylabel={FID cat},
    xticklabels={X, 0, 1M, 2M, 3M, 4M, 5M},
    ymax=50,
    ymajorgrids=true,
    grid style=dashed,
]

\addplot[color=orange]
coordinates {
(0.0,328.62)(240.0,107.82)(480.0,83.24)(720.0,66.36)(960.0,53.11)(1200.0,44.28)(1440.0,39.34)(1680.0,38.87)(1920.0,37.57)(2160.0,34.05)(2400.0,34.31)(2640.0,34.11)(2880.0,32.87)(3120.0,32.61)(3360.0,32.38)(3600.0,32.12)(3840.0,30.07)(4080.0,30.89)(4320.0,30.91)(4560.0,29.31)(4808.0,29.73)(5000.0,30.37)
};

\addplot[color=red]
coordinates {
(0.0,17.16)(240.0,24.55)(480.0,23.68)(720.0,23.98)(960.0,24.56)(1200.0,24.21)(1440.0,23.41)(1680.0,24.28)(1920.0,24.99)(2160.0,25.44)(2400.0,24.20)(2640.0,25.20)(2880.0,24.07)(3120.0,24.09)(3360.0,24.75)(3600.0,23.86)(3840.0,24.42)(4080.0,23.94)(4320.0,24.08)(4560.0,23.64)(4800.0,24.14)(5000.0,23.32)
};

\addplot[color=green]
coordinates {
(0.0,17.16)(240.0,24.17)(480.0,22.85)(720.0,24.65)(960.0,23.33)(1200.0,23.49)(1440.0,23.88)(1680.0,23.30)(1920.0,23.17)(2160.0,24.15)(2400.0,22.88)(2640.0,23.64)(2880.0,23.73)(3120.0,22.73)(3360.0,24.34)(3600.0,23.38)(3840.0,23.51)(4080.0,23.22)(4320.0,23.57)(4560.0,22.99)(4800.0,23.52)(5000.0,23.04)
};

\addplot[color=cyan]
coordinates {
(0.0,17.16)(240.0,23.16)(480.0,22.82)(720.0,24.28)(960.0,23.41)(1200.0,23.70)(1444.0,24.21)(1692.0,24.12)(1932.0,23.31)(2172.0,23.88)(2412.0,23.74)(2652.0,23.78)(2896.0,24.94)(3184.0,26.76)(3428.0,24.26)(3668.0,23.01)(3908.0,23.63)(4148.0,23.53)(4388.0,22.77)(4628.0,23.57)(4868.0,23.60)(5000.0,23.36)
};

\addplot[color=blue]
coordinates {
(0.0,65.29)(240.0,30.98)(480.0,28.23)(720.0,27.15)(960.0,26.41)(1200.0,25.23)(1440.0,24.68)(1680.0,23.70)(1920.0,24.82)(2160.0,24.33)(2400.0,23.71)(2640.0,23.39)(2880.0,23.48)(3120.0,23.14)(3360.0,23.96)(3600.0,23.49)(3840.0,22.45)(4080.0,22.49)(4320.0,22.95)(4560.0,22.13)(4800.0,22.21)(5000.0,22.08)
};

\legend{Scratch, TransferGAN, EWC, Freeze-D, Our}
    
\end{axis}
\end{tikzpicture}

%% file: figures/cat_dog_convergence/dog.tex
\begin{tikzpicture}[scale=0.7]
\begin{axis}[
    xlabel={Number of iterations},
    ylabel={FID dog},
    xticklabels={X, 0, 1M, 2M, 3M, 4M, 5M},
    ymax=50,
    ymajorgrids=true,
    grid style=dashed,
]

\addplot[color=orange]
coordinates {
(0.0,307.73)(240.0,107.18)(480.0,74.50)(720.0,61.17)(960.0,54.02)(1200.0,50.12)(1440.0,44.92)(1680.0,42.20)(1920.0,41.32)(2160.0,38.17)(2400.0,37.77)(2640.0,36.47)(2880.0,35.77)(3120.0,35.75)(3360.0,34.00)(3600.0,34.56)(3840.0,32.50)(4080.0,31.87)(4320.0,33.08)(4560.0,32.15)(4808.0,33.47)(5000.0,33.22)
};

\addplot[color=red]
coordinates {
(0.0,96.93)(240.0,38.88)(480.0,36.02)(720.0,35.91)(960.0,35.06)(1200.0,33.86)(1440.0,33.59)(1680.0,31.92)(1920.0,31.55)(2160.0,31.13)(2400.0,30.68)(2640.0,30.92)(2880.0,31.11)(3120.0,31.04)(3360.0,30.51)(3600.0,31.40)(3840.0,30.03)(4080.0,29.19)(4320.0,29.71)(4560.0,29.98)(4800.0,29.37)(5000.0,28.84)
};

\addplot[color=green]
coordinates {
(0.0,96.93)(240.0,39.32)(480.0,37.85)(720.0,35.55)(960.0,34.31)(1200.0,33.75)(1440.0,32.45)(1680.0,32.02)(1920.0,32.00)(2160.0,31.61)(2400.0,31.26)(2640.0,31.69)(2880.0,31.21)(3120.0,31.56)(3360.0,31.37)(3600.0,31.02)(3840.0,30.60)(4080.0,30.98)(4320.0,29.82)(4560.0,30.17)(4800.0,30.83)(5000.0,30.11)
};

\addplot[color=cyan]
coordinates {
(0.0,96.93)(240.0,38.20)(480.0,35.67)(720.0,34.09)(960.0,32.46)(1200.0,32.09)(1444.0,31.58)(1692.0,30.94)(1932.0,31.85)(2172.0,30.79)(2412.0,30.97)(2652.0,29.82)(2896.0,29.98)(3184.0,30.35)(3428.0,29.74)(3668.0,29.41)(3908.0,29.07)(4148.0,28.23)(4388.0,27.93)(4628.0,28.37)(4868.0,26.88)(5000.0,28.40)
};

\addplot[color=blue]
coordinates {
(0.0,49.13)(240.0,30.35)(480.0,28.64)(720.0,28.72)(960.0,29.04)(1200.0,28.06)(1440.0,28.80)(1680.0,28.05)(1920.0,28.04)(2160.0,27.57)(2400.0,26.85)(2640.0,27.77)(2880.0,27.57)(3120.0,27.06)(3360.0,27.10)(3600.0,27.60)(3840.0,26.86)(4080.0,26.80)(4320.0,26.81)(4560.0,26.40)(4800.0,26.58)(5000.0,26.52)
};

\legend{Scratch, TransferGAN, EWC, Freeze-D, Our}
    
\end{axis}
\end{tikzpicture}

%% file: figures/cat_ffhq_convergence/full.tex
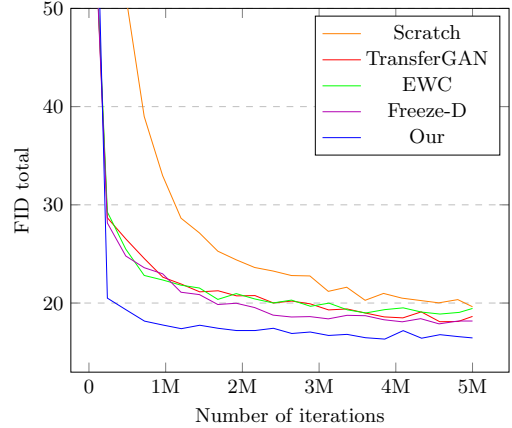
\begin{figure}[t]
\centering
\begin{tikzpicture}[scale=0.85]
\begin{axis}[
    xlabel={Number of iterations},
    ylabel={FID total},
    xticklabels={X, 0, 1M, 2M, 3M, 4M, 5M},
    ymax=50,
    ymajorgrids=true,
    grid style=dashed,
]

\addplot[color=orange]
coordinates {
(0.0,314.02)(240.0,77.79)(480.0,51.32)(720.0,39.00)(960.0,32.98)(1200.0,28.64)(1440.0,27.14)(1680.0,25.28)(1920.0,24.40)(2160.0,23.62)(2400.0,23.26)(2640.0,22.80)(2880.0,22.76)(3120.0,21.19)(3360.0,21.61)(3600.0,20.28)(3840.0,20.98)(4080.0,20.50)(4320.0,20.26)(4560.0,20.02)(4808.0,20.35)(5000.0,19.62)
};

\addplot[color=red]
coordinates {
(0.0,72.27)(240.0,28.68)(484.0,26.50)(724.0,24.51)(964.0,22.63)(1204.0,21.93)(1444.0,21.14)(1684.0,21.25)(1924.0,20.73)(2164.0,20.76)(2404.0,20.01)(2644.0,20.20)(2884.0,19.92)(3124.0,19.30)(3364.0,19.40)(3604.0,18.98)(3848.0,18.59)(4092.0,18.49)(4332.0,19.10)(4572.0,18.10)(4812.0,18.12)(5000.0,18.64)
};

\addplot[color=green]
coordinates {
(0.0,72.14)(240.0,29.27)(480.0,25.49)(720.0,22.81)(960.0,22.35)(1200.0,21.82)(1440.0,21.53)(1680.0,20.37)(1920.0,20.96)(2160.0,20.42)(2400.0,20.01)(2640.0,20.30)(2888.0,19.67)(3128.0,19.99)(3368.0,19.31)(3608.0,18.99)(3856.0,19.33)(4096.0,19.51)(4336.0,19.08)(4576.0,18.88)(4828.0,19.05)(5000.0,19.45)
};

\addplot[color=cyan]
coordinates {
(0.0,72.14)(240.0,28.15)(480.0,24.79)(720.0,23.60)(960.0,22.98)(1200.0,21.10)(1440.0,20.86)(1680.0,19.85)(1920.0,19.98)(2160.0,19.54)(2400.0,18.77)(2640.0,18.58)(2880.0,18.62)(3120.0,18.39)(3364.0,18.75)(3604.0,18.71)(3844.0,18.30)(4084.0,18.09)(4324.0,18.41)(4564.0,17.88)(4812.0,18.17)(5000.0,18.17)
};

\addplot[color=blue]
coordinates {
(0.0,92.88)(240.0,20.50)(484.0,19.31)(724.0,18.16)(964.0,17.76)(1204.0,17.39)(1444.0,17.74)(1684.0,17.43)(1924.0,17.20)(2164.0,17.20)(2404.0,17.43)(2644.0,16.90)(2884.0,17.05)(3124.0,16.70)(3364.0,16.81)(3604.0,16.47)(3856.0,16.33)(4096.0,17.19)(4336.0,16.42)(4576.0,16.78)(4816.0,16.58)(5000.0,16.45)
};

\legend{Scratch, TransferGAN, EWC, Freeze-D, Our}
    
\end{axis}
\end{tikzpicture}
\caption{Convergence rate of the total FID score during training on LSUN cat + FFHQ. Our solution achieves the lowest (best) FID score.}
\label{fig:cat_ffhq_total_fid_convergence_rate}
\end{figure}

%% file: figures/cat_ffhq_convergence/per_class.tex
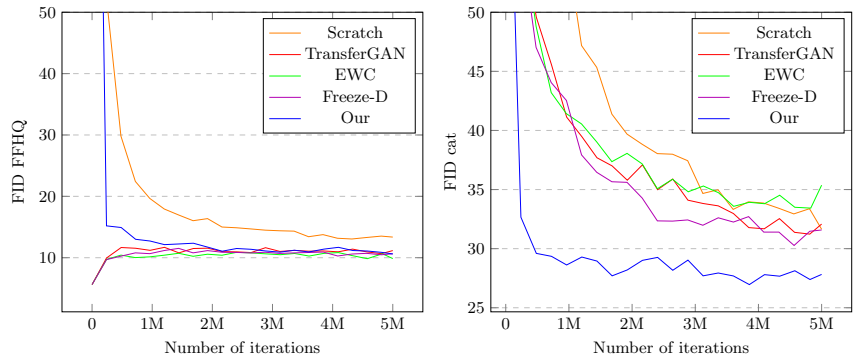
\begin{figure}[t]
    \centering
    \begin{tabular}{cc}
        \input{figures/cat_ffhq_convergence/ffhq}
        & 
        \input{figures/cat_ffhq_convergence/cat}
    \end{tabular}
    \caption{FID scores calculated separately on FFHQ (left) and on cat (right). TransferGAN suffers from catastophic forgetting of FFHQ (left) that is mitigated by EWC which achieves slightly better results on this dataset than our method, but this comes at the expense of the FID score of the cat class (right), which is the worst for EWC out of all methods.}
    \label{fig:cat_ffhq_per_class_fid_convergence_rate}
\end{figure}

%% file: figures/cat_ffhq_convergence/ffhq.tex
\begin{tikzpicture}[scale=0.7]
\begin{axis}[
    xlabel={Number of iterations},
    ylabel={FID FFHQ},
    xticklabels={X, 0, 1M, 2M, 3M, 4M, 5M},
    ymax=50,
    ymajorgrids=true,
    grid style=dashed,
]

\addplot[color=orange]
coordinates {
(0.0,342.09)(240.0,52.35)(480.0,29.82)(720.0,22.43)(960.0,19.66)(1200.0,17.94)(1440.0,16.96)(1680.0,16.03)(1920.0,16.37)(2160.0,15.00)(2400.0,14.88)(2640.0,14.69)(2880.0,14.49)(3120.0,14.39)(3360.0,14.33)(3600.0,13.42)(3840.0,13.76)(4080.0,13.17)(4320.0,13.05)(4560.0,13.29)(4808.0,13.52)(5000.0,13.36)
};

\addplot[color=red]
coordinates {
(0.0,5.62)(240.0,9.95)(484.0,11.66)(724.0,11.54)(964.0,11.20)(1204.0,11.69)(1444.0,10.79)(1684.0,11.50)(1924.0,11.55)(2164.0,10.99)(2404.0,10.95)(2644.0,10.84)(2884.0,11.63)(3124.0,11.02)(3364.0,11.19)(3604.0,11.06)(3848.0,11.10)(4092.0,10.96)(4332.0,11.40)(4572.0,10.94)(4812.0,10.66)(5000.0,11.16)
};

\addplot[color=green]
coordinates {
(0.0,5.63)(240.0,9.72)(480.0,10.42)(720.0,10.02)(960.0,10.14)(1200.0,10.41)(1440.0,10.72)(1680.0,10.23)(1920.0,10.57)(2160.0,10.40)(2400.0,10.87)(2640.0,10.79)(2888.0,10.63)(3128.0,10.49)(3368.0,10.69)(3608.0,10.27)(3856.0,10.71)(4096.0,10.83)(4336.0,10.32)(4576.0,9.84)(4828.0,10.64)(5000.0,9.86)
};

\addplot[color=cyan]
coordinates {
(0.0,5.62)(240.0,9.69)(480.0,10.25)(720.0,10.79)(960.0,10.66)(1200.0,11.17)(1440.0,11.52)(1680.0,10.79)(1920.0,11.16)(2160.0,10.89)(2400.0,10.86)(2640.0,10.78)(2880.0,10.89)(3120.0,10.70)(3364.0,10.79)(3604.0,10.85)(3844.0,10.92)(4084.0,10.28)(4324.0,10.60)(4564.0,10.71)(4812.0,10.48)(5000.0,10.64)
};

\addplot[color=blue]
coordinates {
(0.0,173.82)(240.0,15.19)(484.0,14.92)(724.0,13.00)(964.0,12.72)(1204.0,12.13)(1444.0,12.24)(1684.0,12.36)(1924.0,11.72)(2164.0,11.06)(2404.0,11.50)(2644.0,11.37)(2884.0,11.10)(3124.0,10.91)(3364.0,11.22)(3604.0,10.94)(3856.0,11.40)(4096.0,11.69)(4336.0,11.17)(4576.0,11.09)(4816.0,10.87)(5000.0,10.61)
};

\legend{Scratch, TransferGAN, EWC, Freeze-D, Our}
    
\end{axis}
\end{tikzpicture}

%% file: figures/cat_ffhq_convergence/cat.tex
\begin{tikzpicture}[scale=0.7]
\begin{axis}[
    xlabel={Number of iterations},
    ylabel={FID cat},
    xticklabels={X, 0, 1M, 2M, 3M, 4M, 5M},
    ymax=50,
    ymajorgrids=true,
    grid style=dashed,
]

\addplot[color=orange]
coordinates {
(0.0,314.52)(240.0,118.09)(480.0,84.86)(720.0,65.87)(960.0,55.63)(1200.0,47.18)(1440.0,45.34)(1680.0,41.38)(1920.0,39.69)(2160.0,38.83)(2400.0,38.04)(2640.0,37.99)(2880.0,37.44)(3120.0,34.67)(3360.0,34.98)(3600.0,33.32)(3840.0,33.96)(4080.0,33.86)(4320.0,33.42)(4560.0,32.94)(4808.0,33.38)(5000.0,31.64)
};

\addplot[color=red]
coordinates {
(0.0,193.99)(240.0,57.10)(484.0,49.58)(724.0,45.56)(964.0,41.13)(1204.0,39.48)(1444.0,37.69)(1684.0,37.00)(1924.0,35.79)(2164.0,37.09)(2404.0,34.99)(2644.0,35.89)(2884.0,34.10)(3124.0,33.82)(3364.0,33.62)(3604.0,32.97)(3848.0,31.78)(4092.0,31.68)(4332.0,32.54)(4572.0,31.38)(4812.0,31.23)(5000.0,32.08)
};

\addplot[color=green]
coordinates {
(0.0,194.01)(240.0,58.08)(480.0,48.82)(720.0,43.20)(960.0,41.42)(1200.0,40.55)(1440.0,39.03)(1680.0,37.36)(1920.0,38.06)(2160.0,37.17)(2400.0,35.07)(2640.0,35.87)(2888.0,34.81)(3128.0,35.30)(3368.0,34.73)(3608.0,33.58)(3856.0,33.92)(4096.0,33.80)(4336.0,34.52)(4576.0,33.50)(4828.0,33.43)(5000.0,35.37)
};

\addplot[color=cyan]
coordinates {
(0.0,193.99)(240.0,56.25)(480.0,47.03)(720.0,44.05)(960.0,42.54)(1200.0,37.93)(1440.0,36.46)(1680.0,35.66)(1920.0,35.61)(2160.0,34.28)(2400.0,32.35)(2640.0,32.33)(2880.0,32.43)(3120.0,31.98)(3364.0,32.61)(3604.0,32.25)(3844.0,32.71)(4084.0,31.40)(4324.0,31.40)(4564.0,30.27)(4812.0,31.47)(5000.0,31.58)
};

\addplot[color=blue]
coordinates {
(0.0,76.70)(240.0,32.66)(484.0,29.60)(724.0,29.35)(964.0,28.62)(1204.0,29.29)(1444.0,28.95)(1684.0,27.70)(1924.0,28.20)(2164.0,29.01)(2404.0,29.26)(2644.0,28.17)(2884.0,29.03)(3124.0,27.70)(3364.0,27.94)(3604.0,27.69)(3856.0,26.96)(4096.0,27.80)(4336.0,27.67)(4576.0,28.13)(4816.0,27.39)(5000.0,27.82)
};

\legend{Scratch, TransferGAN, EWC, Freeze-D, Our}
    
\end{axis}
\end{tikzpicture}

%% file: appendices/applications.tex
\section{Applications}
\paragraph{Semantic editing} We demonstrate additional examples for the semantic editing application. We used a merged model of FFHQ and LSUN cat. In \Cref{fig:supp_semantic_editing_pose} we show more examples to
Figure 4 in the main paper: we calculated the human pose direction in the $\mathcal{W}$ latent space of the merged generator on images of FFHQ class only by fitting an SVM that separates images with ``positive'' pose and images with ``negative'' pose, then we used the calculated hyperplane normal and applied it to images that were generated from both of the classes. 
Note how the pose direction also applies to the cats, even though it was calculated using human photos.

\input{figures/supp_semantic_editing_pose/supp_semantic_editing_pose}

In addition, we also experimented with semantic directions whose meaning may be less clear or even undefined for some of the classes.
We did not expect these manipulations to work, but wanted to investigate their behavior.
In \Cref{fig:supp_semantic_editing_gender} we 
calculate the direction in the latent space that corresponds to the gender of the subject on the FFHQ class, and apply this direction to images for both classes. As can be seen, this direction has a clear effect on the FFHQ class, but not on the LSUN cat class, where it mainly affects the size of the cat. Another example can be seen in \Cref{fig:supp_semantic_editing_glasses}, where we calculate the ``add glasses'' direction in the latent space for the FFHQ class. While this direction operates well on the FFHQ class, since the LSUN cat class does not have images of cats with glasses, it is not surprising that the effect is not carried over to cat images. Note, however, that this latent direction does affect the same semantic region --- adding glasses is replaced by slightly increasing the cats' eyes.

\input{figures/supp_semantic_editing_gender/supp_semantic_editing_gender}

\input{figures/supp_semantic_editing_glasses/supp_semantic_editing_glasses}

%% file: figures/supp_semantic_editing_pose/supp_semantic_editing_pose.tex
\begin{figure}[t]
    \centering
    \setlength{\tabcolsep}{0.5pt}
    \renewcommand{\arraystretch}{0.5}
    \setlength{\ww}{0.15\textwidth}
  
    \begin{tabular}{ccccc}
        \includegraphics[width=\ww,frame]{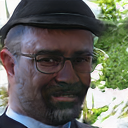} &
        \includegraphics[width=\ww,frame]{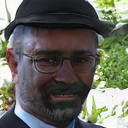} &
        \includegraphics[width=\ww,frame]{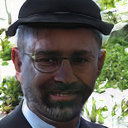} &
        \includegraphics[width=\ww,frame]{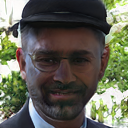} &
        \includegraphics[width=\ww,frame]{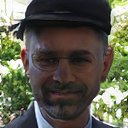} \\
        
        \includegraphics[width=\ww,frame]{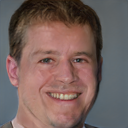} &
        \includegraphics[width=\ww,frame]{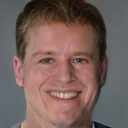} &
        \includegraphics[width=\ww,frame]{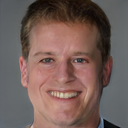} &
        \includegraphics[width=\ww,frame]{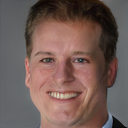} &
        \includegraphics[width=\ww,frame]{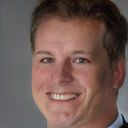} \\
        
        \includegraphics[width=\ww,frame]{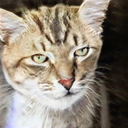} &
        \includegraphics[width=\ww,frame]{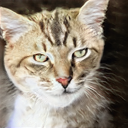} &
        \includegraphics[width=\ww,frame]{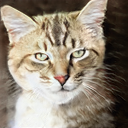} &
        \includegraphics[width=\ww,frame]{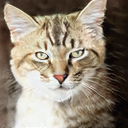} &
        \includegraphics[width=\ww,frame]{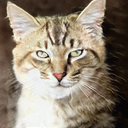} \\
        
        \includegraphics[width=\ww,frame]{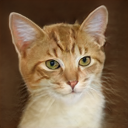} &
        \includegraphics[width=\ww,frame]{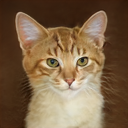} &
        \includegraphics[width=\ww,frame]{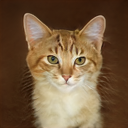} &
        \includegraphics[width=\ww,frame]{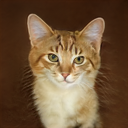} &
        \includegraphics[width=\ww,frame]{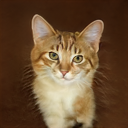} \\
        
        (-) Pose &  & &  & (+) Pose
    \end{tabular}
    
    \caption{We determine the pose direction in the latent space of the merged model of FFHQ and LSUN cat using images of the FFHQ class only. We then apply this direction to images from both classes and find that the semantics are largely preserved.}
    \label{fig:supp_semantic_editing_pose}
\end{figure}

%% file: figures/supp_semantic_editing_gender/supp_semantic_editing_gender.tex
\begin{figure}[t]
    \centering
    \setlength{\tabcolsep}{0.5pt}
    \renewcommand{\arraystretch}{0.5}
    \setlength{\ww}{0.15\textwidth}
  
    \begin{tabular}{ccccc}
        \includegraphics[width=\ww,frame]{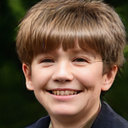} &
        \includegraphics[width=\ww,frame]{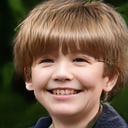} &
        \includegraphics[width=\ww,frame]{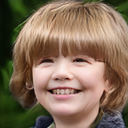} &
        \includegraphics[width=\ww,frame]{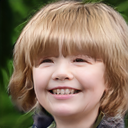} &
        \includegraphics[width=\ww,frame]{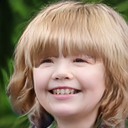} \\
        
        \includegraphics[width=\ww,frame]{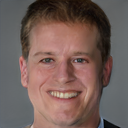} &
        \includegraphics[width=\ww,frame]{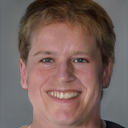} &
        \includegraphics[width=\ww,frame]{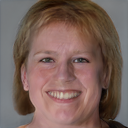} &
        \includegraphics[width=\ww,frame]{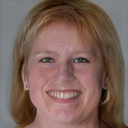} &
        \includegraphics[width=\ww,frame]{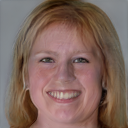} \\
        
        \includegraphics[width=\ww,frame]{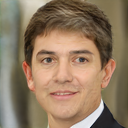} &
        \includegraphics[width=\ww,frame]{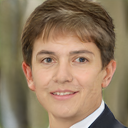} &
        \includegraphics[width=\ww,frame]{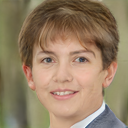} &
        \includegraphics[width=\ww,frame]{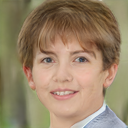} &
        \includegraphics[width=\ww,frame]{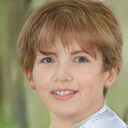} \\
        
        \includegraphics[width=\ww,frame]{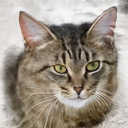} &
        \includegraphics[width=\ww,frame]{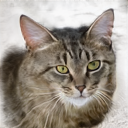} &
        \includegraphics[width=\ww,frame]{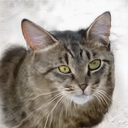} &
        \includegraphics[width=\ww,frame]{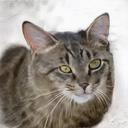} &
        \includegraphics[width=\ww,frame]{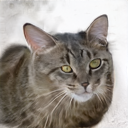} \\
        
        \includegraphics[width=\ww,frame]{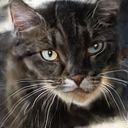} &
        \includegraphics[width=\ww,frame]{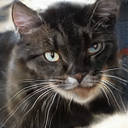} &
        \includegraphics[width=\ww,frame]{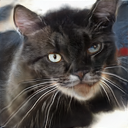} &
        \includegraphics[width=\ww,frame]{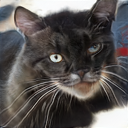} &
        \includegraphics[width=\ww,frame]{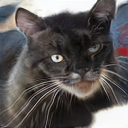} \\
        
        \includegraphics[width=\ww,frame]{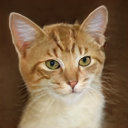} &
        \includegraphics[width=\ww,frame]{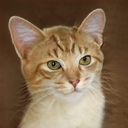} &
        \includegraphics[width=\ww,frame]{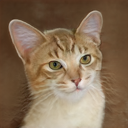} &
        \includegraphics[width=\ww,frame]{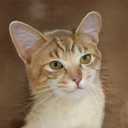} &
        \includegraphics[width=\ww,frame]{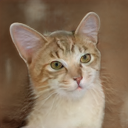} \\
        
        Male &  &  &  & Female
    \end{tabular}
    
    \caption{We determine the gender direction in the latent space of the merged model of FFHQ and LSUN cat using images of the FFHQ class only. We then apply this direction to images from both classes. As expected, this operates accurately only on the FFHQ class.}
    \label{fig:supp_semantic_editing_gender}
\end{figure}

%% file: figures/supp_semantic_editing_glasses/supp_semantic_editing_glasses.tex
\begin{figure}[t]
    \centering
    \setlength{\tabcolsep}{0.5pt}
    \renewcommand{\arraystretch}{0.5}
    \setlength{\ww}{0.15\textwidth}
  
    \begin{tabular}{ccccc}
        \includegraphics[width=\ww,frame]{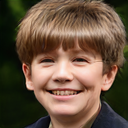} &
        \includegraphics[width=\ww,frame]{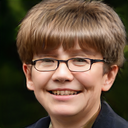} &
        \includegraphics[width=\ww,frame]{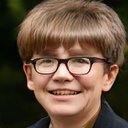} &
        \includegraphics[width=\ww,frame]{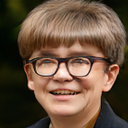} &
        \includegraphics[width=\ww,frame]{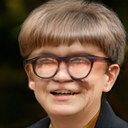} \\
        
        \includegraphics[width=\ww,frame]{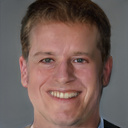} &
        \includegraphics[width=\ww,frame]{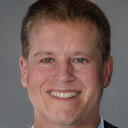} &
        \includegraphics[width=\ww,frame]{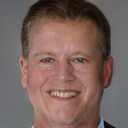} &
        \includegraphics[width=\ww,frame]{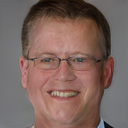} &
        \includegraphics[width=\ww,frame]{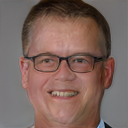} \\
        
        \includegraphics[width=\ww,frame]{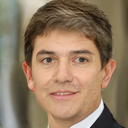} &
        \includegraphics[width=\ww,frame]{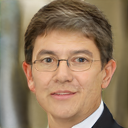} &
        \includegraphics[width=\ww,frame]{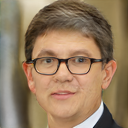} &
        \includegraphics[width=\ww,frame]{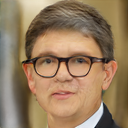} &
        \includegraphics[width=\ww,frame]{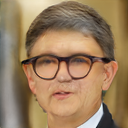} \\
    
        \includegraphics[width=\ww,frame]{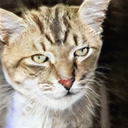} &
        \includegraphics[width=\ww,frame]{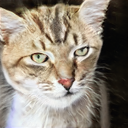} &
        \includegraphics[width=\ww,frame]{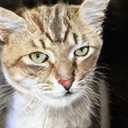} &
        \includegraphics[width=\ww,frame]{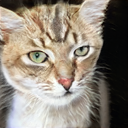} &
        \includegraphics[width=\ww,frame]{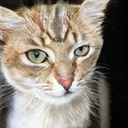} \\
        
        \includegraphics[width=\ww,frame]{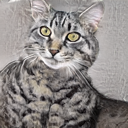} &
        \includegraphics[width=\ww,frame]{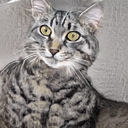} &
        \includegraphics[width=\ww,frame]{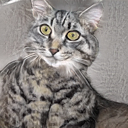} &
        \includegraphics[width=\ww,frame]{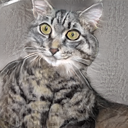} &
        \includegraphics[width=\ww,frame]{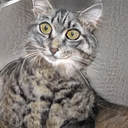} \\
        
        \includegraphics[width=\ww,frame]{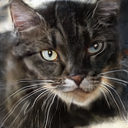} &
        \includegraphics[width=\ww,frame]{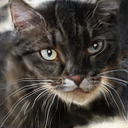} &
        \includegraphics[width=\ww,frame]{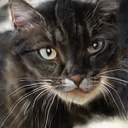} &
        \includegraphics[width=\ww,frame]{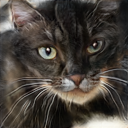} &
        \includegraphics[width=\ww,frame]{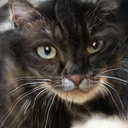} \\
        
        \includegraphics[width=\ww,frame]{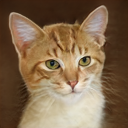} &
        \includegraphics[width=\ww,frame]{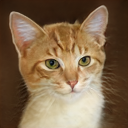} &
        \includegraphics[width=\ww,frame]{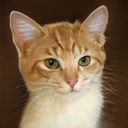} &
        \includegraphics[width=\ww,frame]{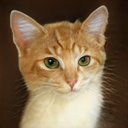} &
        \includegraphics[width=\ww,frame]{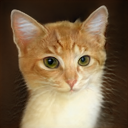} \\
        
        No glasses &  & &  & Add glasses
    \end{tabular}
    
    \caption{We determine the glasses direction in the latent space of the merged model of FFHQ and LSUN cat using images of the FFHQ class only. We then apply this direction to images from both classes. The addition of glasses operates accurately only on the FFHQ class (as expected). On the cat class the same direction enlarges the eyes of the cats.}
    \label{fig:supp_semantic_editing_glasses}
\end{figure}

%% file: appendices/uncurated_examples.tex
\section{Uncurated Generation Examples}

In \Cref{fig:samples_dog} and \Cref{fig:samples_cat} we present uncurated images generated by the input source GAN models, by the baselines, and by our method.

\input{figures/uncurated_samples/dog/fig.tex}
\input{figures/uncurated_samples/cat/fig.tex}

%% file: figures/uncurated_samples/dog/fig.tex
\begin{figure}[ht]
    \centering
    \begin{tabular}{cc}
        \subfloat[Source model]{\includegraphics[width=0.45\textwidth]{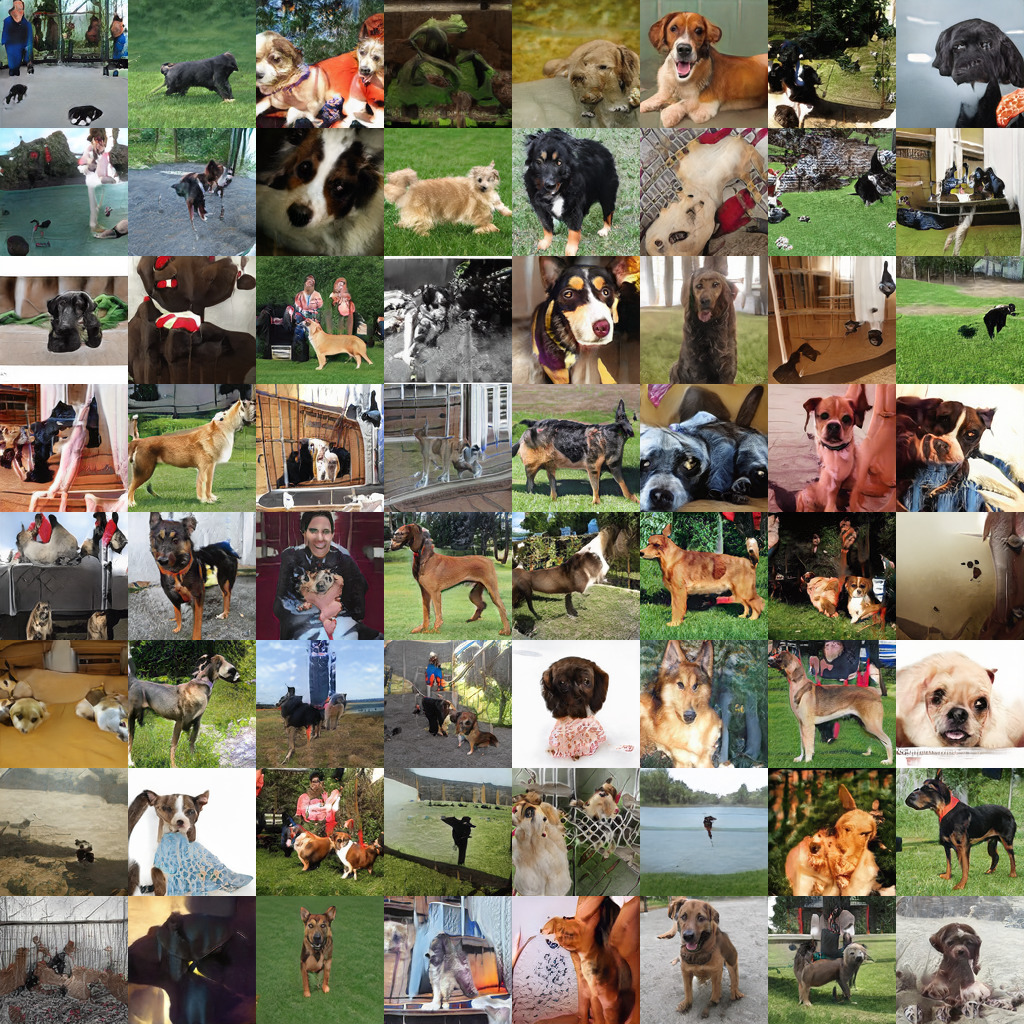}} &
        \subfloat[From scratch]{\includegraphics[width=0.45\textwidth]{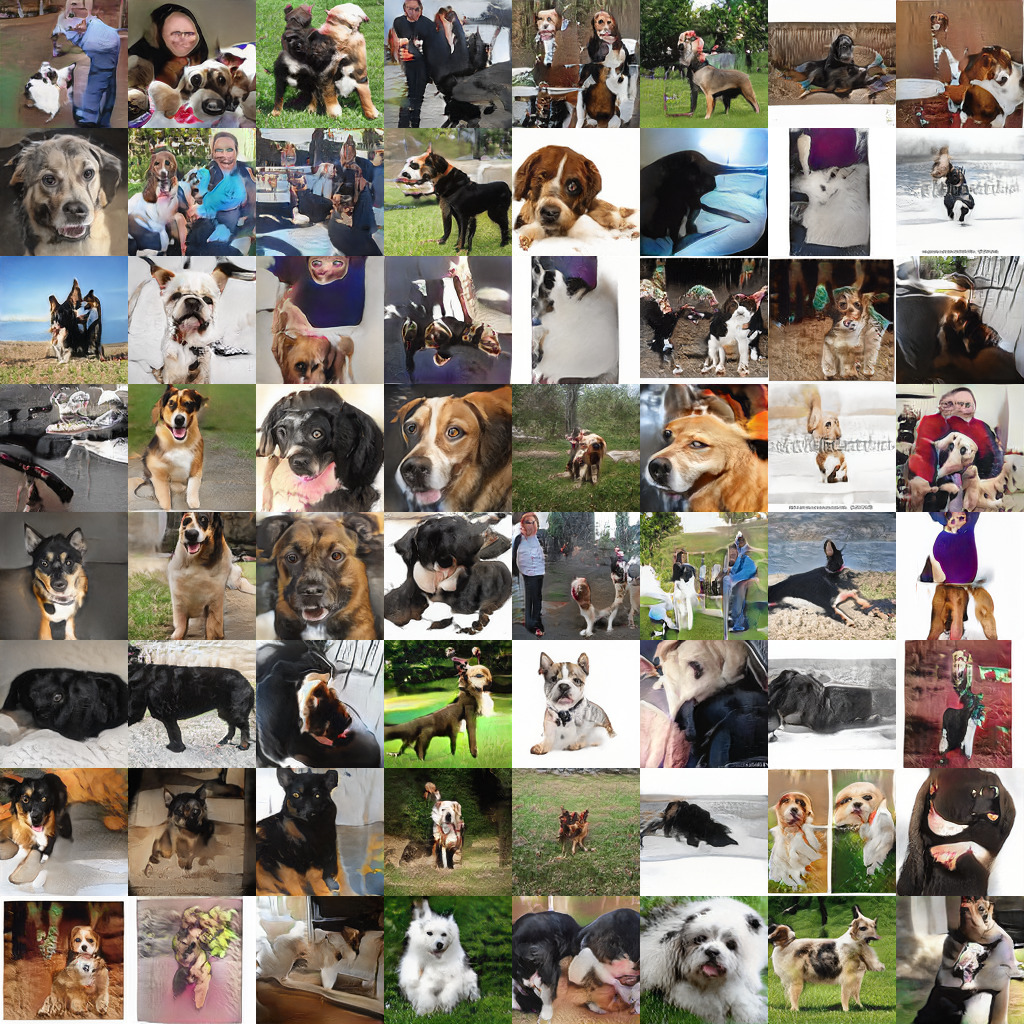}}
        \\

        \subfloat[TransferGAN]{\includegraphics[width=0.45\textwidth]{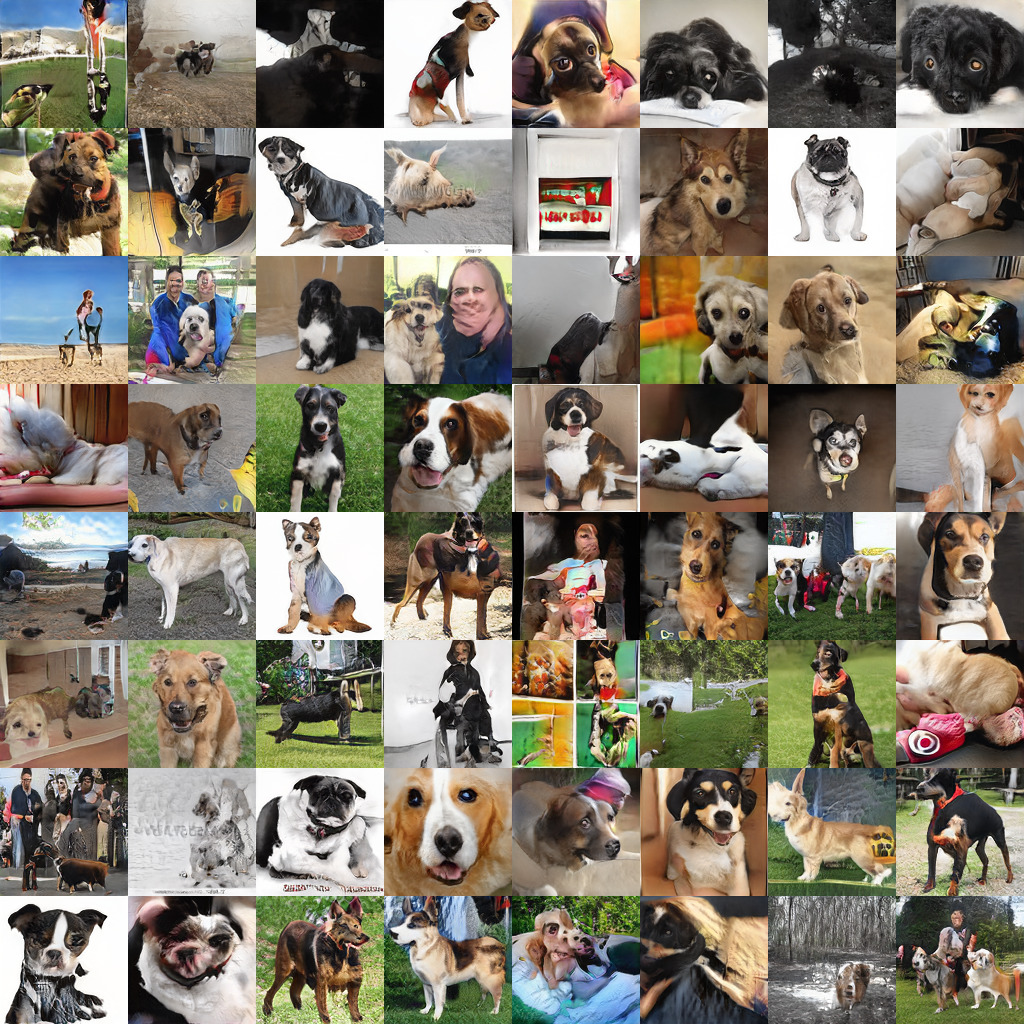}} &
        \subfloat[FreezeD]{\includegraphics[width=0.45\textwidth]{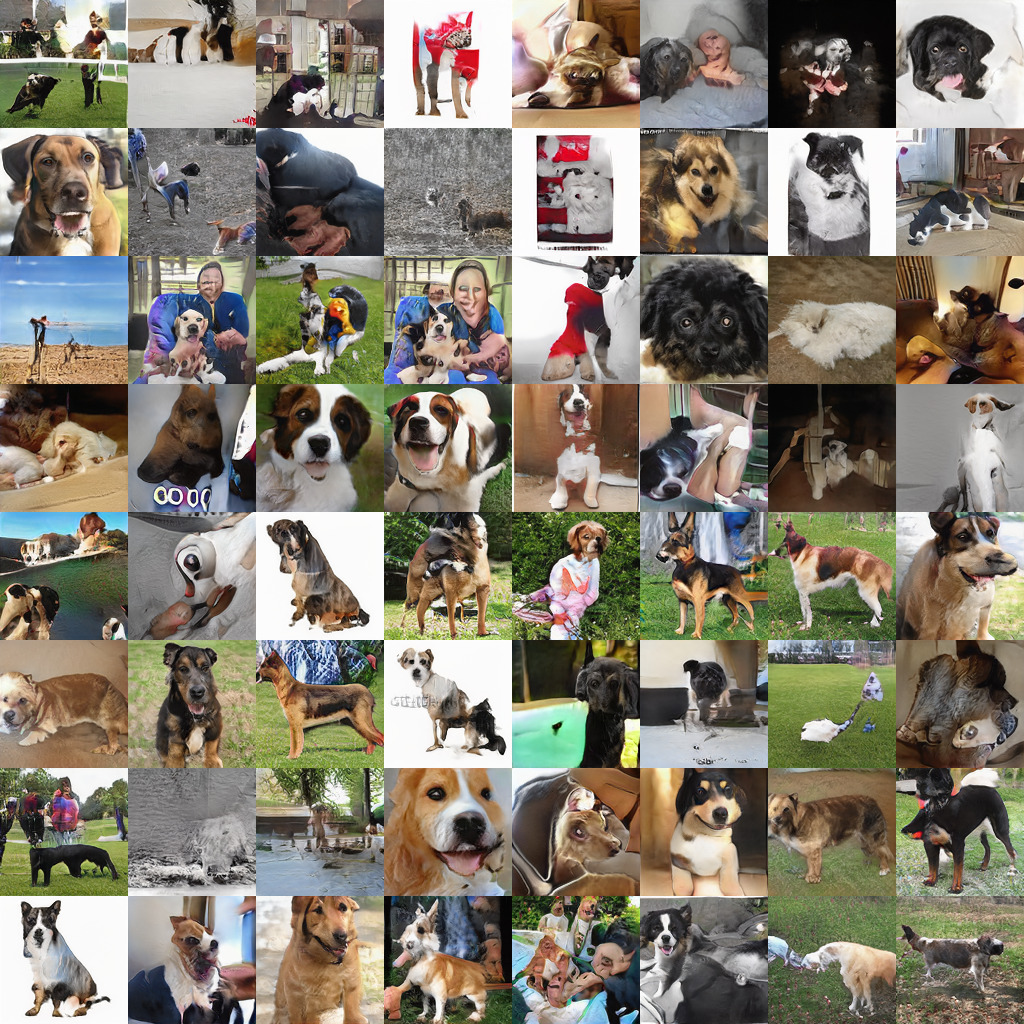}}
        \\

        \subfloat[EWC]{\includegraphics[width=0.45\textwidth]{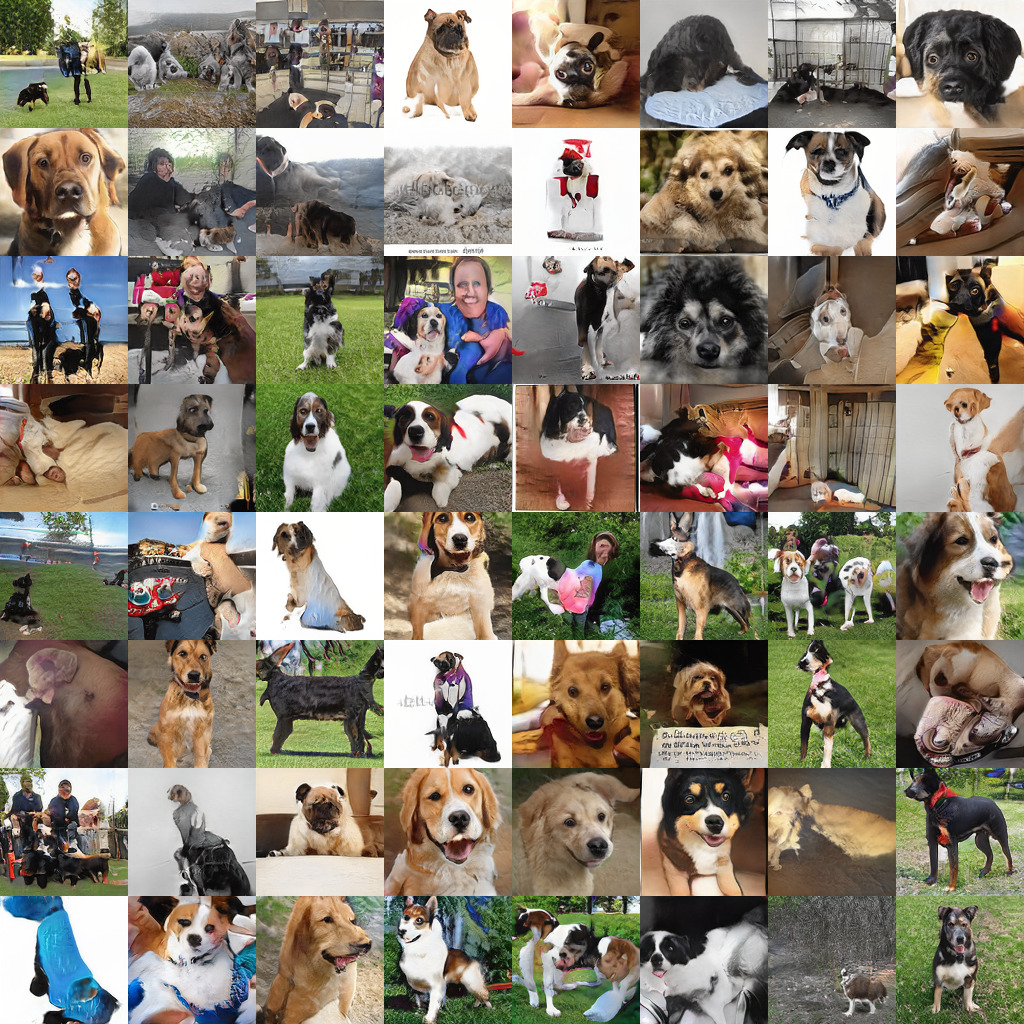}} &
        \subfloat[Ours]{\includegraphics[width=0.45\textwidth]{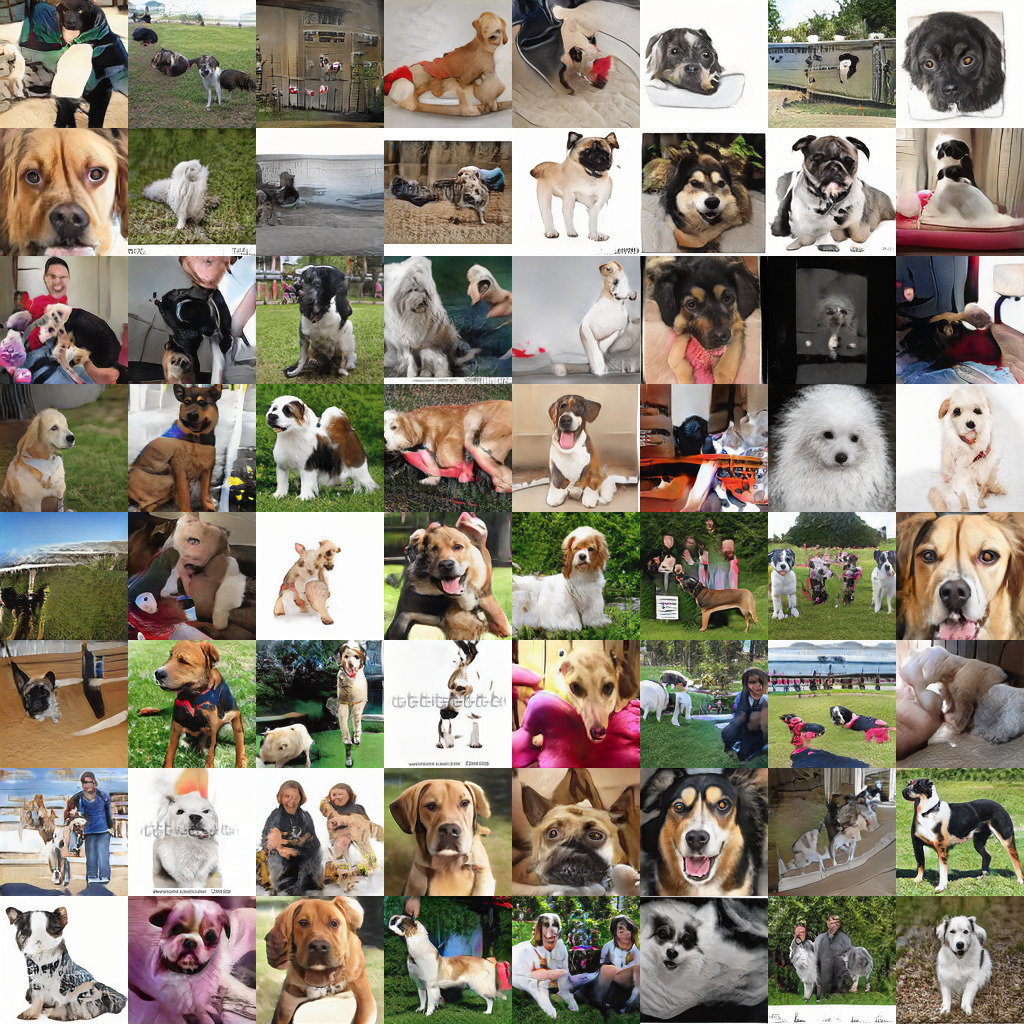}}
        \\
    \end{tabular}
    \caption{Examples of uncurated images that were generated by the source model (a), the baselines (b-e), and our method (f)  on LSUN dog dataset.}
    \label{fig:samples_dog}
\end{figure}

%% file: figures/uncurated_samples/cat/fig.tex
\begin{figure}[ht]
    \centering
    \begin{tabular}{cc}
        \subfloat[Source model]{\includegraphics[width=0.45\textwidth]{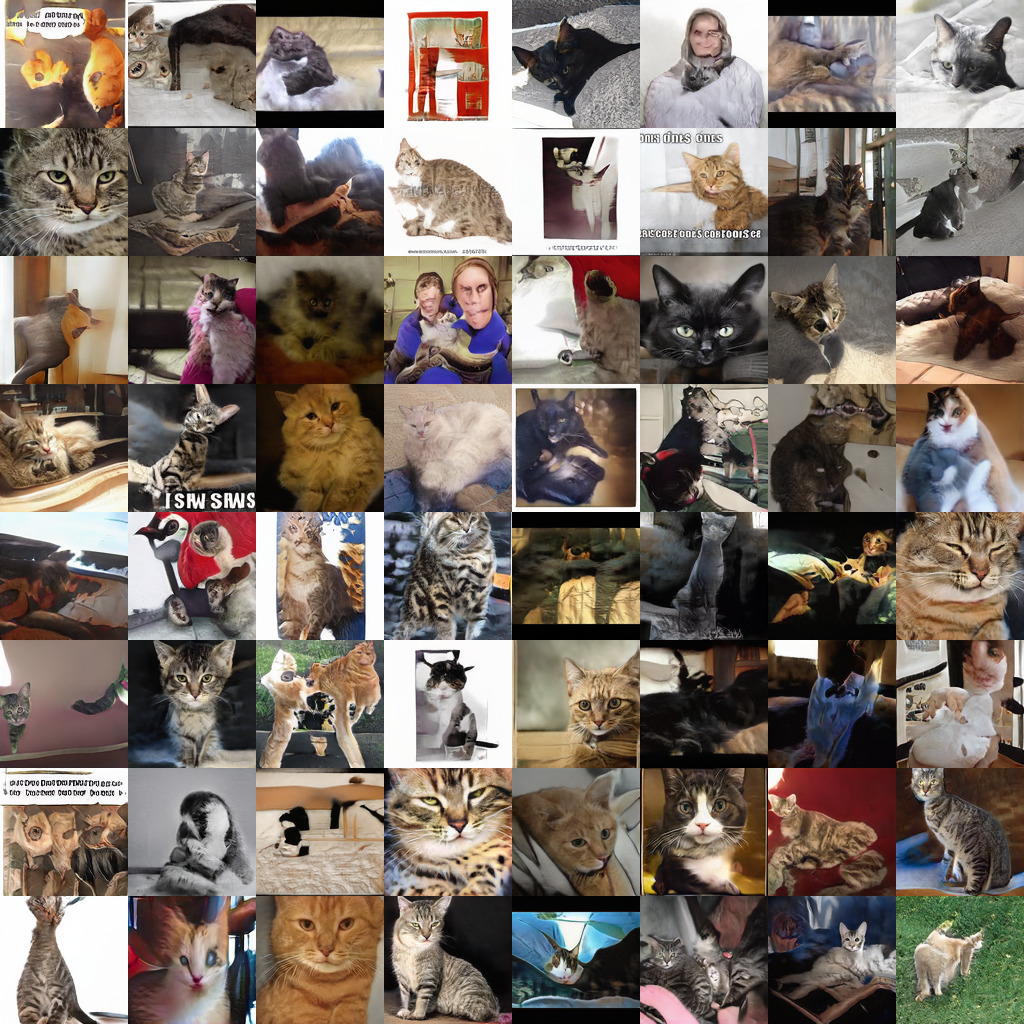}} &
        \subfloat[From scratch]{\includegraphics[width=0.45\textwidth]{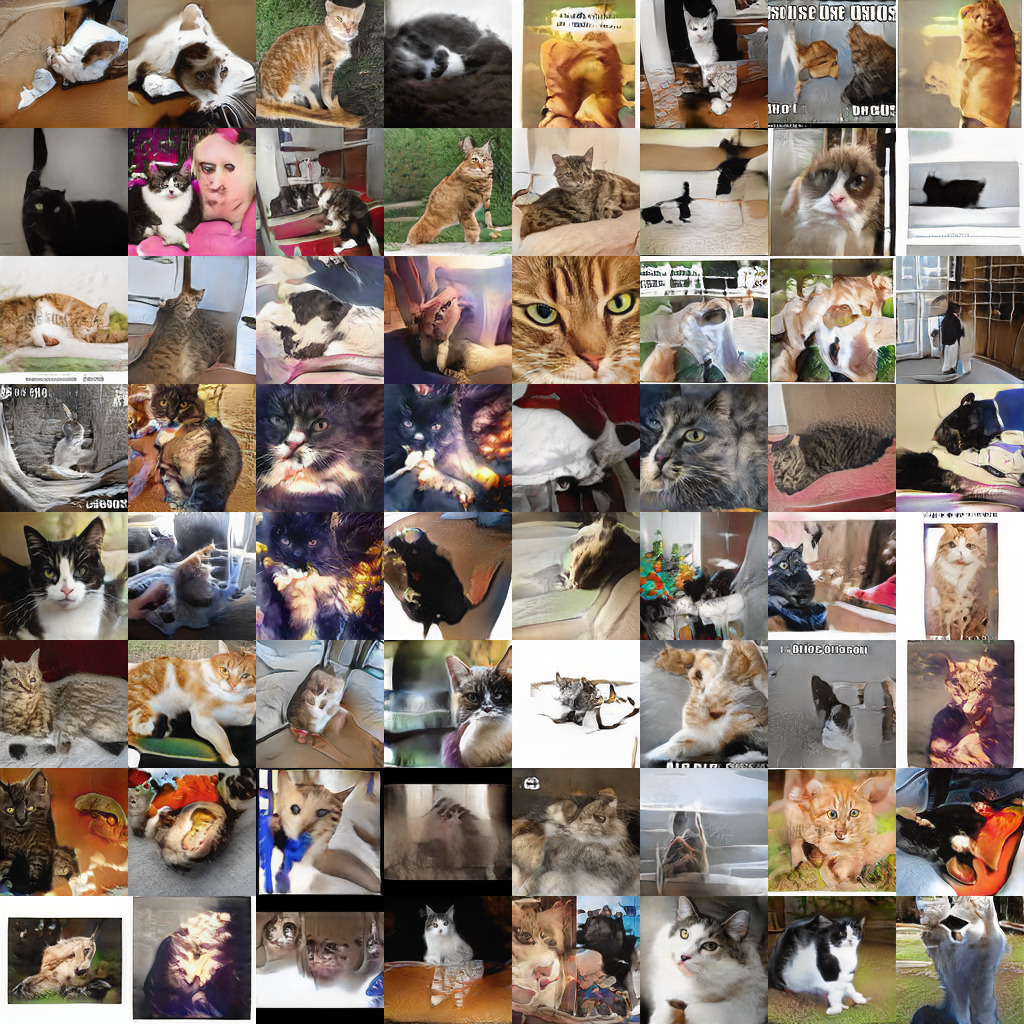}}
        \\

        \subfloat[TransferGAN]{\includegraphics[width=0.45\textwidth]{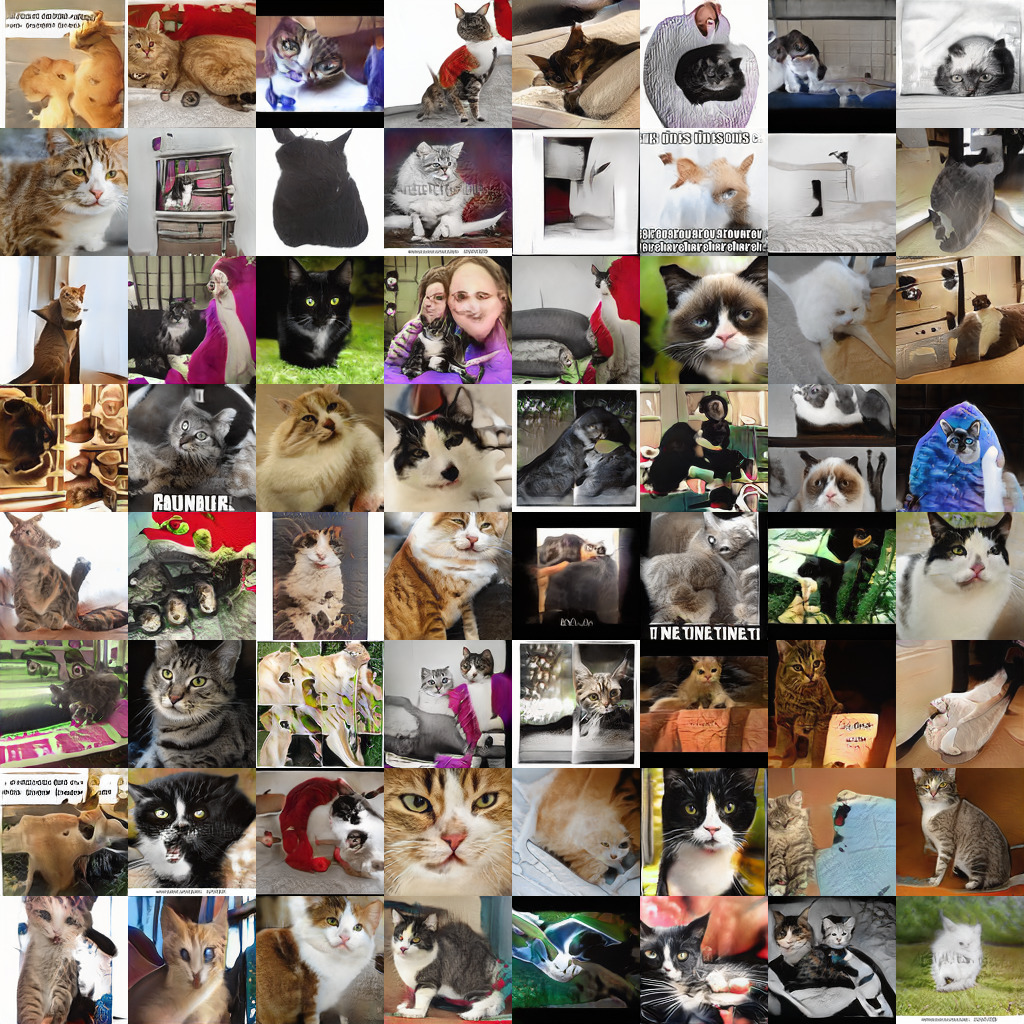}} &
        \subfloat[FreezeD]{\includegraphics[width=0.45\textwidth]{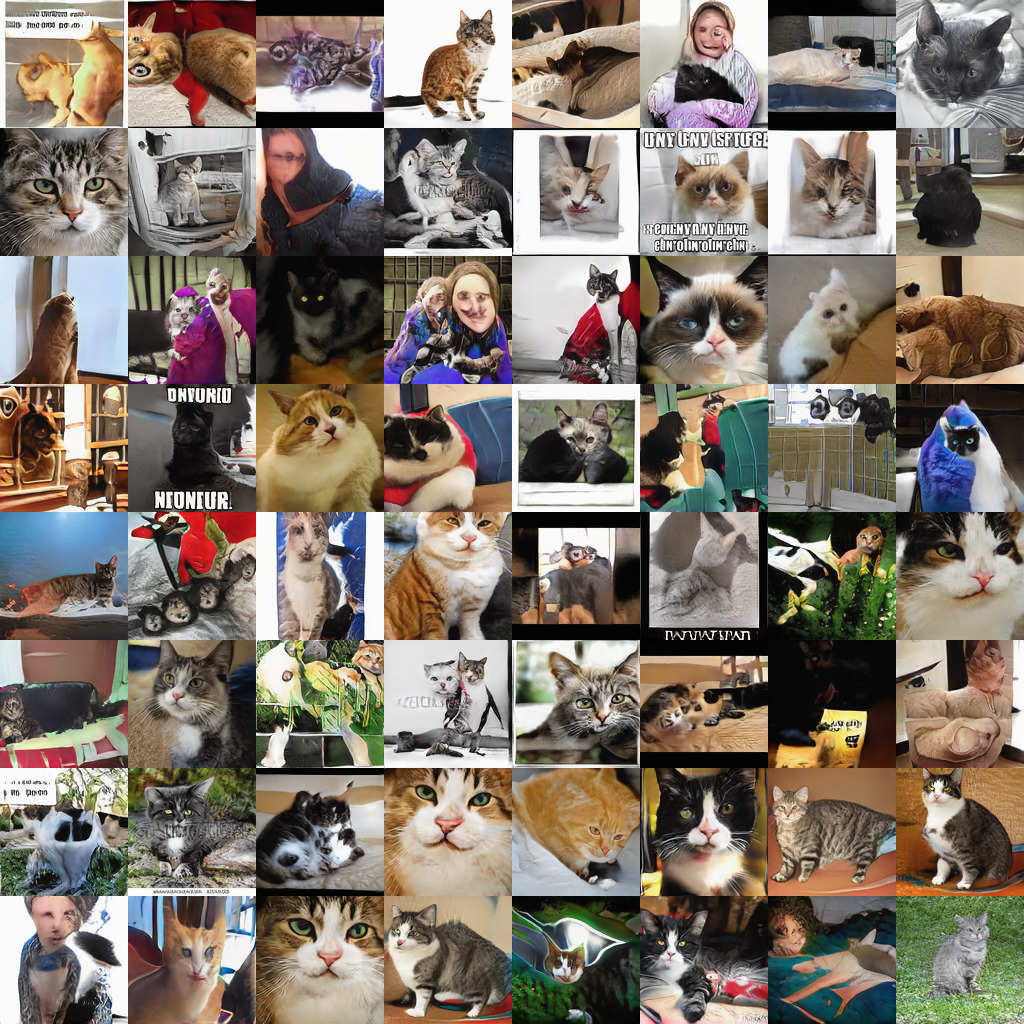}}
        \\

        \subfloat[EWC]{\includegraphics[width=0.45\textwidth]{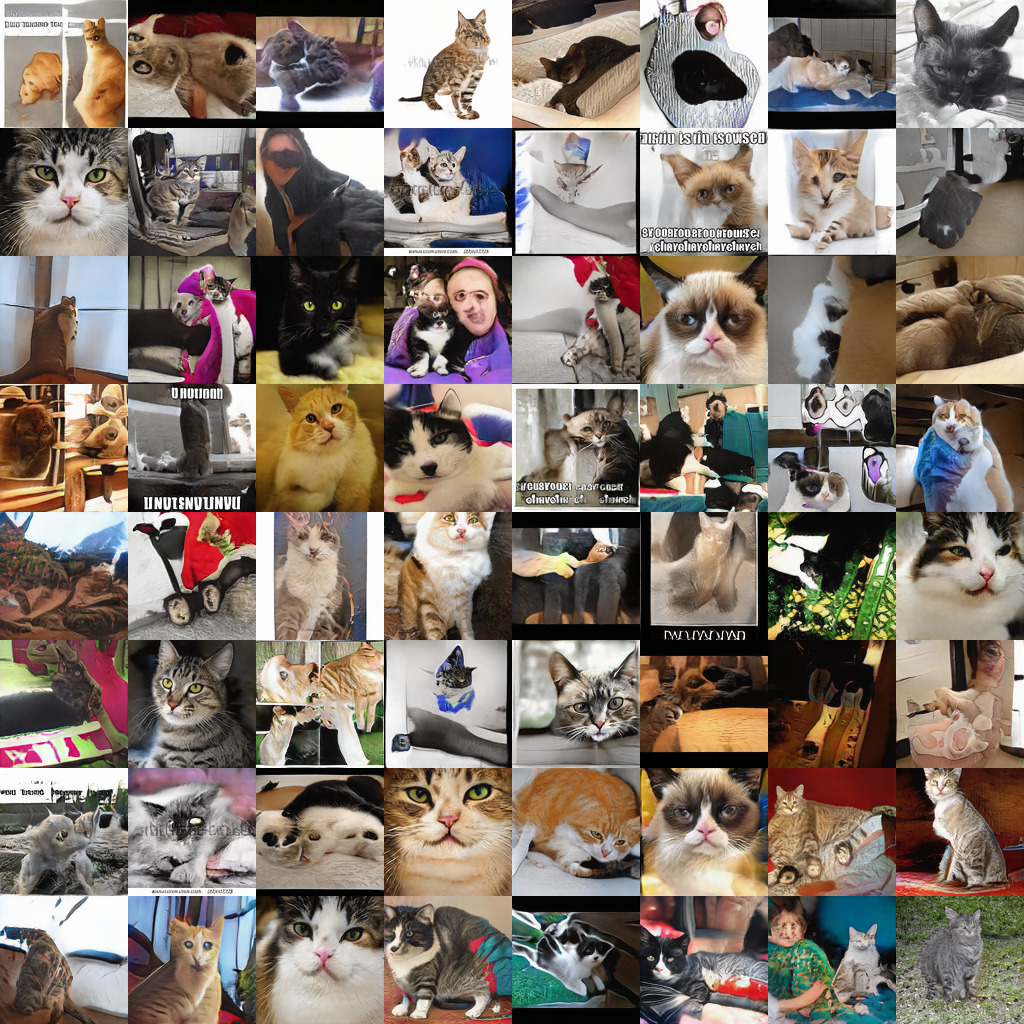}} &
        \subfloat[Ours]{\includegraphics[width=0.45\textwidth]{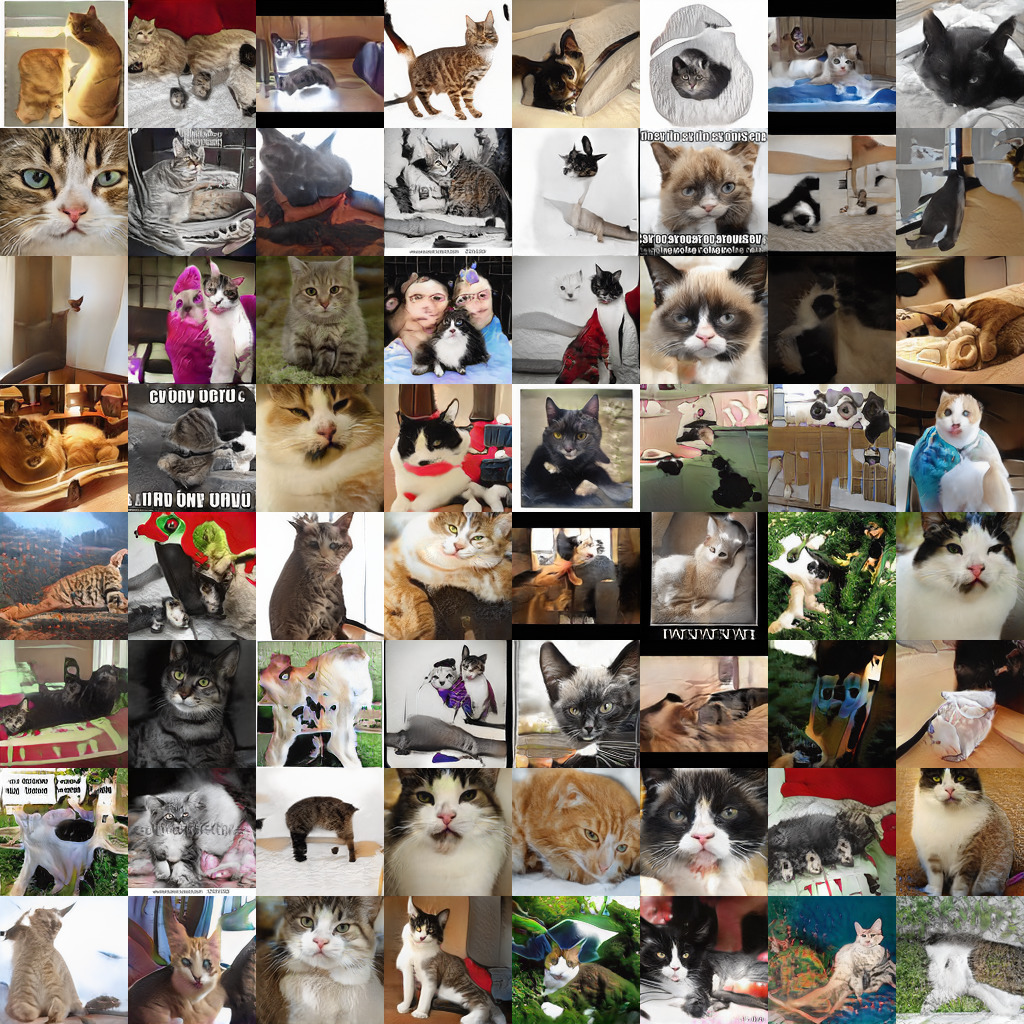}}
        \\
    \end{tabular}
    \caption{Examples of uncurated images that were generated by the source model (a), the baselines (b-e), and our method (f)  on LSUN cat dataset.}
    \label{fig:samples_cat}
\end{figure}